%% file: icml_main.tex
\documentclass{article}

\usepackage{microtype}
\usepackage{graphicx}
\usepackage{booktabs} 

\usepackage{hyperref}

\usepackage[accepted]{icml2023}

\usepackage{amsmath}
\usepackage{amssymb}
\usepackage{mathtools}
\usepackage{amsthm}

\usepackage{url}            
\usepackage{amsfonts}       
\usepackage{nicefrac}       
\usepackage{xcolor}         
\usepackage{amsmath, amsthm, amssymb}
\usepackage{enumitem}
\usepackage{graphicx}
\usepackage{subcaption}

\input{macros}

\usepackage[capitalize,noabbrev]{cleveref}

\usepackage[textsize=tiny]{todonotes}

\icmltitlerunning{Privacy-Aware Compression for Federated Learning}

\begin{document}

\twocolumn[
\icmltitle{Privacy-Aware Compression for Federated Learning Through\\Numerical Mechanism Design}





\icmlsetsymbol{equal}{*}

\begin{icmlauthorlist}
\icmlauthor{Chuan Guo}{meta}
\icmlauthor{Kamalika Chaudhuri}{meta}
\icmlauthor{Pierre Stock}{meta}
\icmlauthor{Mike Rabbat}{meta}
\end{icmlauthorlist}

\icmlaffiliation{meta}{Meta AI}

\icmlcorrespondingauthor{Chuan Guo}{chuanguo@meta.com}
\icmlcorrespondingauthor{Kamalika Chaudhuri}{kamalika@meta.com}

\icmlkeywords{Machine Learning, ICML}

\vskip 0.3in
]



\printAffiliationsAndNotice{} 



\begin{abstract}
In private federated learning (FL), a server aggregates differentially private updates from a large number of clients in order to train a machine learning model. The main challenge in this setting is balancing privacy with both classification accuracy of the learnt model as well as the number of bits communicated between the clients and server. Prior work has achieved a good trade-off by designing a privacy-aware compression mechanism, called the minimum variance unbiased (MVU) mechanism, that numerically solves an optimization problem to determine the parameters of the mechanism. This paper builds upon it by introducing a new interpolation procedure in the numerical design process that allows for a far more efficient privacy analysis. The result is the new Interpolated MVU mechanism that is more scalable, has a better privacy-utility trade-off, and provides SOTA results on communication-efficient private FL on a variety of datasets.
\end{abstract}

\input{intro}
\input{background}
\input{method}
\input{experiment}
\input{relatedwork}
\input{conclusion}

\bibliography{citations}
\bibliographystyle{icml2023}

\newpage
\appendix
\onecolumn
\setcounter{lemma}{0}
\setcounter{theorem}{0}
\input{appendix}

\end{document}

%% file: macros.tex
\newtheorem{lemma}{Lemma}
\newtheorem{theorem}{Theorem}

\newtheorem{definition}{Definition}

\newcommand{\bu}{\mathbf{u}}

\newcommand{\be}{\mathbf{e}}
\newcommand{\bx}{\mathbf{x}}

\newcommand{\bz}{\mathbf{z}}

\newcommand{\ba}{\mathbf{a}}
\newcommand{\bp}{\mathbf{p}}
\newcommand{\bseta}{\boldsymbol{\eta}}
\newcommand{\bstheta}{\boldsymbol{\theta}}

\newcommand{\calD}{\mathcal{D}}

\newcommand{\calM}{\mathcal{M}}

\newcommand{\calI}{\mathcal{I}}

\newcommand{\var}{\text{Var}}

\DeclareMathOperator*{\argmin}{arg\,min}
\DeclareMathOperator*{\argmax}{arg\,max}

%% file: intro.tex
\section{Introduction}

Federated learning (FL; \citet{mcmahan2017communication}) is a distributed solution for machine learning on sensitive data and has been the topic of much recent interest. In federated learning, raw client data resides entirely on the client device, and a server coordinates the collaborative training of a global model through transmissions of model updates.
Direct transmission of raw updates (pseudo-gradients) can lead to privacy leaks~\citep{zhu2019deep, geiping2020inverting, zhao2020idlg, yin2021see, jeon2021gradient}, and an effective method is to use a rigorous privacy solution such as differential privacy (DP; \citet{dwork2006calibrating}) to transmit privacy-preserving updates.
Since federated learning typically involves low-bandwidth clients, a requirement for success is that the amount of communication between the clients and server should be relatively low. Thus a major challenge in this space is ensuring a good privacy-accuracy-communication trade-off---in other words, balancing the predictive accuracy of the model learnt with the privacy parameter as well as the number of bits transmitted between the client and server.

Previous work in this area~\citep{kairouz2021distributed, agarwal2021skellam, triastcyn2021dp, chen2022poisson} abstracted this problem as privacy-aware compression for distributed mean estimation (DME). Here, the goal is to design a DP mechanism whose output recovers the input in expectation (\emph{i.e.}, is unbiased) and can be represented using a small number of bits.
In particular, recent work~\cite{chaudhuri2022privacy} introduces the \emph{minimum variance unbiased} (MVU) mechanism, which solves a numerical optimization problem to find the parameters of a mechanism that ensures unbiasedness, privacy and communication efficiency while minimizing the variance of the output. They also experimentally demonstrate that, for a given bit budget, this can lead to better privacy-utility trade-offs than many other methods that independently privatize and then compress. 

Unfortunately, a problem with the MVU mechanism (and many other privacy-aware compression mechanisms such as \citet{kairouz2021distributed, agarwal2021skellam}) is that it does not scale well to high-dimensional vectors. This is a crucial drawback for its application to private FL since model updates are often high-dimensional. A major underlying reason for this drawback is the MVU mechanism's insistence on unbiasedness. This is achieved using \emph{randomized dithering} to ensure that the mechanism is unbiased for all continuous-valued inputs, but leads to inefficiencies in the mechanism's computation and privacy accounting for high-dimensional vectors. Importantly, this drawback is in fact superficial since unbiasedness is not required for FL, and prior work has shown that even crude compressors such as SignSGD~\citep{jin2020stochastic} can perform remarkably well.

In this work, we propose the \emph{interpolated MVU} (I-MVU) mechanism, which extends MVU using an interpolation procedure that relaxes the unbiasedness assumption. By its discrete nature, the MVU mechanism can be viewed as sampling from a particular categorical distribution, and hence can be expressed in exponential family form. The proposed I-MVU mechanism handles continuous-valued inputs by \emph{interpolating the natural exponential family parameters}, rather than directly interpolating the probabilities as in dithering. We introduce a new analysis technique, and by further exploiting special properties of the exponential family, we obtain a tight \emph{numerical} privacy analysis for the vector extension under both $L_1$ and $L_2$ geometry. 

Experimentally, we find that under both client-level and sample-level DP settings and across various benchmark datasets, the I-MVU mechanism provides a better privacy-utility trade-off than SignSGD~\cite{jin2020stochastic} and MVU~\cite{chaudhuri2022privacy} at an extremely low communication budget of \emph{one bit per gradient dimension}. Moreover, I-MVU achieves close to the same performance as the standard non-compressed Laplace and Gaussian mechanisms~\citep{abadi2016deep} for similar levels of ($\epsilon, \delta)$-DP, leading to new state-of-the-art results for private communication-efficient FL.

%% file: background.tex
\section{Background}

\subsection{Differential Privacy}


Differential privacy~\cite{dwork2006calibrating} is a cryptographically-motivated definition of privacy that has emerged as the gold standard in privacy-preserving data analysis, and provides rigorous guarantees of privacy leakage induced by a mechanism $\calM$. 

\begin{definition}[Differential privacy] We say that a mechanism $\calM$ that applies to datasets $\calD$ is $(\epsilon,\delta)$-differentially private, denoted $(\epsilon,\delta)$-DP, if for any two neighboring datasets $\calD$ and $\calD'$ that differ\footnote{We adopt the leave-one-out neighboring notion.} in a single individual's private value, and any output set $O$, we have:
\begin{equation*}
    \mathbb{P}(\calM(\calD) \in O) \leq e^{\epsilon} \mathbb{P}(\calM(\calD') \in O) + \delta.
\end{equation*}
\end{definition}

More generally, the framework of DP seeks to bound the difference in distribution between $\calM(\calD)$ and $\calM(\calD')$ when $\calD$ and $\calD'$ differ by a single record that corresponds to one person's private value. Enforcing this property ensures that a single person's data does not substantially change the probability of any event that comes out of using the output of the mechanism $\calM$.

\paragraph{R\'{e}nyi DP and composition.} A useful variant of DP is R\'{e}nyi differential privacy (RDP)~\cite{mironov2017renyi}, which instead bounds the R\'{e}nyi divergence~\cite{renyi1961measures} between $\calM(\calD)$ and $\calM(\calD')$ by some $\epsilon$ for neighboring datasets $\calD$ and $\calD'$ that differ by a single person's private value. Formally, we say that $\calM$ is $(\alpha, \epsilon)$-RDP if:
\begin{equation*}
	\max(D_\alpha(\calM(\calD)~||~\calM(\calD')), D_\alpha(\calM(\calD')~||~\calM(\calD))) \leq \epsilon,
\end{equation*}
where $D_\alpha$ denotes the order-$\alpha$ R\'{e}nyi divergence.

An important property of R\'{e}nyi DP is that it supports easy ``sequential composition'' of mechanisms---in other words, it makes it relatively simple to measure how the privacy guarantee decays as we release the outputs of multiple differentially private mechanisms on the same dataset $\calD$. Specifically, we have that if
 $\calM_1,\ldots,\calM_T$ are mechanisms with $\calM_t$ being $(\alpha, \epsilon_t)$-RDP for $t=1,\ldots,T$, then the composition of the $T$ mechanisms is $(\alpha, \sum_{t=1}^T \epsilon_t)$-RDP. Another useful property of RDP is its conversion to $(\epsilon,\delta)$-DP~\cite{balle2020hypothesis}: If $\calM$ is $(\alpha, \epsilon_\alpha)$-RDP for $\alpha > 1$ then it is also $(\epsilon,\delta)$-DP for any $0 < \delta < 1$ with
\begin{equation}
    \label{eq:rdp_conversion}
    \epsilon = \epsilon_\alpha + \log \left( \frac{\alpha-1}{\alpha} \right) - \frac{\log \delta + \log \alpha}{\alpha - 1}.
\end{equation}

\subsection{Federated Learning with Differential Privacy} 

Federated learning (FL)~\cite{mcmahan2017communication,kairouz2021advances} allows distributed training of ML models across multiple clients without centralized data storage. A server coordinates training by acquiring model updates from clients, aggregating them, and then transmitting an updated model back to the clients, and the process repeats until convergence. One promise of FL is data privacy since the updates are computed locally on each client using their own data, and hence no client data is ever explicitly transmitted to the server (or anyone else) throughout training.

In spite of this, a recent line of work showed that it is possible to reconstruct training samples from the model updates in a process called \emph{gradient inversion}~\cite{zhu2019deep, geiping2020inverting, zhao2020idlg}.
To truly preserve the privacy of training data in FL, differential privacy (DP; \citet{dwork2006calibrating} can be applied to the model updates~\citep{geyer2017differentially} to provide provable guarantees against data reconstruction from the its output~\cite{balle2022reconstructing, guo2022bounding, stock2022defending}. To apply DP to FL training, given a DP mechanism $\calM$ and a client update $\bx$, the client instead sends $\calM(\bx)$ to the server. For a given round, the client's privacy leakage can be computed in terms of \emph{local DP} if the privatized update $\calM(\bx)$ is revealed to the server, or \emph{global DP} if secure aggregation~\citep{bonawitz2017practical} is applied to aggregate the privatized updates before revealing it to the server. The total privacy leakage throughout training can then be computed via RDP composition and conversion to $(\epsilon,\delta)$-DP via \autoref{eq:rdp_conversion}.


\subsection{MVU Mechanism}

\paragraph{Privacy-aware compression.} Since model updates in FL are high-dimensional vectors of size equal to the number of model parameters, it is also important in practice to compress these updates for communication efficiency.
Prior work studied this \emph{privacy-aware compression} problem under the abstraction of distributed mean estimation (DME), where clients aim to compute the mean of a set of vectors in a distributed fashion under communication constraint.
Formally stated, consider the problem of transmitting a vector $\bx \in \mathbb{R}^d$ differentially privately using at most $bd$ bits, with $b \ge 1$ small enough so that the entire vector $\bx$ can be transmitted efficiently. One can reduce this problem to a scalar one by considering how to privately compress $x \in [0,1]$ using at most $b$ bits, and then scaling the vector $\bx$ appropriately and applying the scalar mechanism coordinate-wise.

\paragraph{Scalar MVU.} The \emph{minimum variance unbiased} (MVU) mechanism~\cite{chaudhuri2022privacy} solves the privacy-aware compression problem by first discretizing the interval $[0,1]$ into $B_\text{in}$ points $\mathcal{X} = \{x_1=0, x_2,\ldots,x_{B_\text{in}}=1\}$ with $x_i := (i-1)/(B_\text{in}-1)$. If $x = x_i$, the mechanism samples $j \sim \text{Categorical}(\bp_i)$ using a probability vector $\bp_i \in \Delta^{B_\text{out}-1}$ and outputs $\calM(x) = a_j \in \mathbb{R}$ where $\{a_1,\ldots,a_{B_\text{out}}\}$ is a pre-determined output alphabet. The probability vectors $\bp_1,\ldots,\bp_{B_\text{in}}$ and output alphabet $\{a_1,\ldots,a_{B_\text{out}}\}$ are numerically designed so that the mechanism satisfies the following three properties:
\begin{enumerate}[noitemsep, leftmargin=*, topsep=0pt]
    \item \emph{$\epsilon$-differential privacy}: $e^{-\epsilon} \bp_{i',j} \leq \bp_{i,j} \leq e^\epsilon \bp_{i',j}$ for all $i \neq i'$ and all $j$.
    \item \emph{Unbiasedness}: $\sum_{j=1}^{B_\text{out}} a_j \bp_{i,j} = x_i$ for all $i$.
    \item \emph{Minimum variance}: $\sum_{i=1}^{B_\text{in}} \var(\calM(x_i))$ is minimal among all mechanisms satisfying 1 and 2.
\end{enumerate}
The MVU mechanism can then be applied to all $x \in [0,1]$ by \emph{randomly dithering} $x$ to the nearest $x_i$ and $x_{i+1}$ such that the dithering is unbiased in expectation, \emph{i.e.},
\begin{equation*}
\mathcal{M}(x) \sim
\begin{cases}
\text{Categorical}(\bp_i) & \text{w.p. } (x_{i+1}-x)/\Delta, \\
\text{Categorical}(\bp_{i+1}) & \text{w.p. } (x - x_i)/\Delta,
\end{cases}
\end{equation*}
where $\Delta = 1/(B_\text{in}-1)$.
One can also view this dithering procedure as linearly interpolating between $\bp_i$ and $\bp_{i+1}$: $\mathcal{M}(x) \sim \text{Categorical}(\bp(x))$ with
\begin{equation}
    \label{eq:mvu_interpolation}
     \bp(x) = \left( \frac{x_{i+1} - x}{\Delta} \right) \bp_i + \left( \frac{x - x_i}{\Delta} \right) \bp_{i+1}.
\end{equation}
It is straightforward to generalize the mechanism to any bounded $x$ by scaling it to $[0,1]$ and then applying the MVU mechanism.

\paragraph{Vector MVU.} For a $d$-dimensional vector $\bx$, the MVU mechanism can be applied independently to each coordinate and the privacy cost is $d\epsilon$ by composition if $\bx \in [0,1]^d$ (or in general, if $\| \bx \|_\infty$ is bounded). The conversion from scalar to vector MVU for privatizing FL updates is more complex since DP training in FL commonly require bounds on $\| \bx \|_1$ or $\| \bx \|_2$ instead of $\| \bx \|_\infty$~\cite{geyer2017differentially}. To handle these settings, \citet{chaudhuri2022privacy} proposed an alternative design that replaced property 1 with a metric-DP~\citep{chatzikokolakis2013broadening} variant: Given a metric $d(x,x')$, the mechanism satisfies
\begin{equation*}
\epsilon\textit{-metric-DP: } e^{-\epsilon d(x_i, x_{i'})} \bp_{i',j} \leq \bp_{i,j} \leq e^{\epsilon d(x_i, x_{i'})} \bp_{i',j}
\end{equation*}
for all $i \neq i'$ and all $j$. Designing MVU with metric-DP property with $L_1$ (resp. $L_2$) metric allows it to compose more gracefully across dimensions for $L_1$-bounded (resp. $L_2$-bounded) vectors.

%% file: method.tex
\begin{figure*}[t]
\centering
\includegraphics[width=\textwidth]{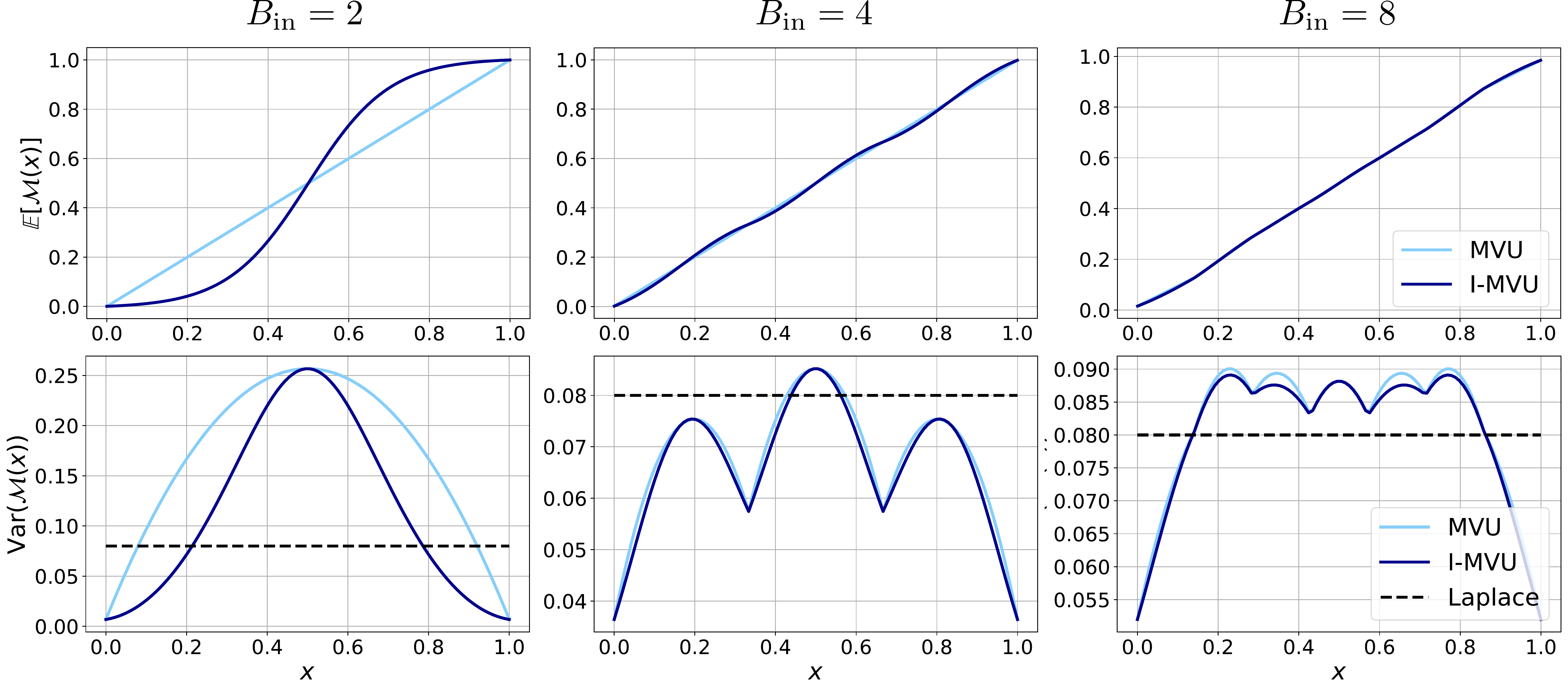}
\caption{Comparison of MVU and I-MVU with $b=3$ and $L_1$-metric-DP $\epsilon=5$. (\textbf{Top}) Plot of the expected value of MVU and I-MVU mechanisms for a scalar input $x \in [0,1]$. For $B_\text{in}=2$, the expected value of MVU is close to a diagonal line with zero bias, but the I-MVU mechanism incurs a significant bias due to interpolating in the natural parameter space. Increasing the input grid size to $B_\text{in}=4,8$ reduces the bias of I-MVU drastically to near zero. (\textbf{Bottom}) Plot of the variance of MVU and I-MVU in comparison to that of the Laplace mechanism with equal $\epsilon$. Both mechanisms achieve comparable variance to the Laplace mechanism when $B_\text{in}$ is large enough.}
\label{fig:imvu_bias_var}
\end{figure*}

\section{Interpolated MVU Mechanism}


\paragraph{Motivation.} At a high level, distributed mean estimation in general and its application to FL have different considerations and requirements.
\begin{enumerate}[noitemsep, leftmargin=*, topsep=0pt]
\item The unbiasedness requirement in DME is often unnecessary for FL, as the global model does not require a truly unbiased gradient in order to achieve high accuracy.
\item Model updates in FL are typically high-dimensional, which creates issues for the randomized dithering operation used in MVU and other privacy-aware compression mechanisms~\cite{kairouz2021distributed, agarwal2021skellam}. For vectors $\bx \in \mathbb{R}^d$, dithering increases $\| \bx \|_1$ and $\| \bx \|_2$ significantly when $d$ is large, hence rejection sampling is needed to control the norm of $\bx$ after dithering. This makes it computationally expensive to apply MVU and causing its privacy accounting to be pessimistic.
\end{enumerate}

\paragraph{Interpolated MVU.} To address these issues, we propose a new interpolation scheme for extending the MVU mechanism to continuous-valued vector inputs. Our new scheme relaxes the unbiasedness requirement in MVU and makes it more suitable for FL use cases. Our first insight is that the categorical distribution used in MVU can be expressed in \emph{exponential family form} as follows:
\begin{equation}
    \label{eq:natural_param}
    \mathbb{P}(j | \bseta) = \exp(\be_j^\top \bseta - A(\bseta)), \; A(\bseta) = \log \left( \sum\nolimits_j \exp(\bseta_j) \right),
\end{equation}
where $\be_j$ is the $j$-th standard basis vector. Note that if $\bp \in \Delta^{B_\text{out}-1}$ then its natural parameter is $\bseta = \log \bp$.

Expressing the MVU mechanism in exponential family form gives us an alternative way to interpolate the MVU mechanism across the grid points $\mathcal{X} = \{x_1,\ldots,x_{B_\text{in}}\}$: by interpolating linearly in the \emph{natural parameter} space.
In detail, consider first the scalar setting with input $x \in [0,1]$. Let $\bp_1, \ldots, \bp_{B_\text{in}} \in \Delta^{B_\text{out}-1}$ be sampling probability vectors obtained from the MVU mechanism and let $\bseta_i = \log \bp_i$ be the natural parameters. Given $x \in [0,1]$, say $x \in [x_i, x_{i+1}]$ for some $i$, the \emph{interpolated MVU} (I-MVU) mechanism samples $j \sim \mathbb{P}(\cdot | \bseta(x))$ according to \autoref{eq:natural_param} and outputs $a_j$ from the MVU output alphabet, where
\begin{equation}
    \label{eq:imvu_interpolation}
     \bseta(x) = \left( \frac{x_{i+1} - x}{\Delta} \right) \bseta_i + \left( \frac{x - x_i}{\Delta} \right) \bseta_{i+1}
\end{equation}
and $\Delta = 1/(B_\text{in} - 1)$.
In other words, instead of linearly interpolating between $\bp_i$ and $\bp_{i+1}$ as in \autoref{eq:mvu_interpolation} for the MVU mechanism, we linearly interpolate between the natural parameters $\bseta_i$ and $\bseta_{i+1}$.

\paragraph{Controlling bias.} For MVU, because the mechanism is designed so that it is unbiased at all grid points in $\mathcal{X}$, it is easy to show via \autoref{eq:mvu_interpolation} and linearity of expectation that it is unbiased for all $x \in [0,1]$. This is not true for I-MVU, and interpolating in the natural parameter space incurs some bias when $x \notin \mathcal{X}$. However, by increasing the number of grid points $B_\text{in}$, we can reduce this bias to arbitrarily small with very few drawbacks.

\autoref{fig:imvu_bias_var} shows the expectation (top) and variance (bottom) of MVU and I-MVU for different values of $B_\text{in}=2,4,8$. The mechanisms are constructed to output $b=3$ bits with target $L_1$-metric-DP $\epsilon=5$. For $B_\text{in}=2$, we see that $\mathbb{E}[\mathcal{M}(x)]=x$ for all $x \in [0,1]$ for MVU, while the same does not hold for I-MVU. As we increase the number of grid points to $B_\text{in}=4,8$, the expectation of I-MVU more closely matches that of MVU and the bias tends towards zero.
In the bottom plots we show the variance of MVU and I-MVU for different $x \in [0,1]$ and compare it with that of the Laplace mechanism for the same value of $\epsilon=5$. Notably, for $B_\text{in}=4,8$, MVU and I-MVU attain comparable variance to that of Laplace while communicating only $b=3$ bits.


\subsection{Extensions}

\paragraph{Vector I-MVU.} We can extend scalar I-MVU to vector I-MVU by applying the scalar mechanism independently across the vector coordinates. As we will show in \autoref{sec:analysis}, by utilizing special properties of exponential family distributions, we can more tightly analyze the privacy cost of I-MVU for $L_1$- and $L_2$-bounded vectors. In addition, because I-MVU does not need to use rejection sampling to control the gradient norm after randomized dithering, privatizing the input simply amounts to sampling a single categorical random variable per coordinate and is much more computationally efficient.

\textbf{Input scaling.} One way to extend I-MVU to arbitrary bounded ranges is to first scale the input to $[0,1]$ and then apply the mechanism as usual. However, note that the interpolation scheme in \autoref{eq:imvu_interpolation} is in fact well-defined for any $x \in \mathbb{R}$, and hence the scaled input does not need to be strictly in the range $[0,1]$. We leverage this property by introducing a scaling factor $\beta \ge 0$ as follows. For $u \in [-C,C]$, the $\beta$-scaled I-MVU mechanism is defined as:
\begin{equation*}
    \calM_\beta(u) = \calM \left( \frac{1}{2} + \frac{\beta u}{2C} \right),
\end{equation*}
where $\calM$ is the plain I-MVU mechanism. Note that this scaling effectively ensures that the input to $\calM$ is in the range $\left[ (1-\beta)/2, (1+\beta)/2 \right]$, with $\beta=1$ corresponding to scaling the input to $[0,1]$. For vectors $\bu$ with $\| \bu \|_2 \leq C$, the $\beta$-scaled input $\bx = \frac{1}{2} + \frac{\beta \bu}{2C}$ satisfies $\| \bx \|_2 \leq \beta/2$.

One advantage for using $\beta$-scaling is that if the distribution of $u$ is highly concentrated near zero, then scaling with $\beta>1$ ensures that the input to $\calM$ is more spread out in the range $[0,1]$. Because the MVU mechanism is designed to minimize the average variance across the range $[0,1]$ (property 3), doing so allows us to better utilize its minimum variance property. For $L_1$- and $L_2$-bounded vectors $\bu$ this is especially true, where the distribution of coordinates of $\bu$ is likely concentrated near zero, hence $\beta$-scaling with a large $\beta$ is essential for achieving good performance.

\subsection{Privacy Analysis}
\label{sec:analysis}

We analyze privacy leakage of the I-MVU mechanism in terms of both pure DP and R\'{e}nyi DP~\cite{mironov2017renyi}. We first present a general analysis for the I-MVU mechanism with any $B_\text{in} \ge 1$ applied to $L_1$-bounded vectors in Theorem~\ref{thm:dp_vector}. Then we give a more specialized analysis for I-MVU with $B_\text{in}=2$ applied to $L_2$-bounded vectors in Theorem~\ref{thm:rdp_vector}. Proofs of theorems and lemmas are given in Appendix \ref{sec:proofs}.

\paragraph{$L_1$-bounded vector analysis.} Our strategy is to first analyze the scalar mechanism and express its max divergence for two differing inputs $x$ and $x'$ as a function of $|x' - x|$ (Lemma \ref{lem:dp_scalar}). Then, by independently applying the mechanism across coordinates, we can sum the max divergence across coordinates and upper bound the total DP $\epsilon$ as a function of $\| \bx' - \bx \|_1$ (Theorem \ref{thm:dp_vector}).

\begin{lemma}
\label{lem:dp_scalar}
Suppose $x, x' \in [0, 1]$ and the MVU mechanism satisfies $\epsilon$ $L_1$-metric-DP. Let $\epsilon' = (B_\text{in} - 1) \times (\max_i \max_{\xi \in [x_i, x_{i+1}]} |\sigma(\bseta(\xi))^\top (\bseta_{i+1} - \bseta_i)|)$. Then:
\begin{equation*}
    D_\infty(\mathbb{P}(\cdot|\bseta(x)) ~||~ \mathbb{P}(\cdot|\bseta(x'))) \leq (\epsilon + \epsilon') |x' - x|.
\end{equation*}
\end{lemma}

\begin{theorem}
\label{thm:dp_vector}
Let $\epsilon'$ be the constant from Lemma \ref{lem:dp_scalar}. Suppose the MVU mechanism is $\epsilon$ $L_1$-metric-DP and that $\bx, \bx' \in \mathbb{R}^d$ satisfy $\| \bx' - \bx \|_1 \leq C$, then the I-MVU mechanism is $(\epsilon + \epsilon') C$-DP.
\end{theorem}

The constant $\epsilon'$ can be computed numerically by maximizing the function $h(\xi) = |\sigma(\bseta(\xi))^\top (\bseta_{i+1} - \bseta_i)|$ over the range $\xi \in [x_i, x_{i+1}]$ for every $i$.

\paragraph{$L_2$-bounded vector analysis.} Our privacy analysis for $L_2$-bounded vectors follows a similar strategy, but instead depends on a measure of information known as \emph{Fisher information}, which we define below for completeness. Intuitively, Fisher information measures the curvature of the R\'{e}nyi divergence, \emph{i.e.}, the rate of change of divergence with respect to input $x$.

\begin{definition}
\label{def:fisher_info}
Let $f$ be the density function of a distribution parameterized by $x \in \mathbb{R}$. The Fisher information of $x$ contained in a sample $Z \sim f(\cdot | x)$ is:
\begin{equation}
    \calI_Z(x) := \mathbb{E}_Z \left[ \left( \frac{d}{dx} \log f(Z | x) \right)^2 \right].
\end{equation}
\end{definition}

In our setting, the distribution $\mathbb{P}(\cdot | \bseta(x))$ is defined by the private data $x$, and Fisher information measures how much information is revealed about $x$ through a sample $j \sim \mathbb{P}(\cdot | \bseta(x))$. It is noteworthy that such a reasoning has also been used to define Fisher information as a privacy metric~\cite{hannun2021measuring}.

For Fisher information to be well-defined, it is necessary that the distribution has a differentiable log density function. For the I-MVU interpolation scheme this is only true if $B_\text{in}=2$; otherwise, the log density may be non-differentiable when transitioning from an interval $[x_{i-1}, x_i]$ to $[x_i, x_{i+1}]$. Thus, Lemma \ref{lem:rdp_scalar} and Theorem \ref{thm:rdp_vector} below are specialized to $B_\text{in}=2$, which we find is sufficient for most FL applications with $L_2$-bounded vectors.

\begin{lemma}
\label{lem:rdp_scalar}
Let $B_\text{in}=2$ and let $M = \sup_{x \in \mathbb{R}} \calI_Z(x)$ be an upper bound on the Fisher information of the mechanism $\calM$. Then for any $x_1, x_2 \in \mathbb{R}$:
\begin{equation}
    \label{eq:rdp_bound}
    D_\alpha(\mathbb{P}(\cdot|\bseta(x_1)) ~||~ \mathbb{P}(\cdot|\bseta(x_2))) \leq \alpha M (x_2 - x_1)^2 / 2.
\end{equation}
\end{lemma}

Applying Lemma \ref{lem:rdp_scalar} in a similar manner as in the $L_1$-bounded case by summing the R\'{e}nyi divergence across coordinates as a function of $\| \bx' - \bx \|_2^2$ yields Theorem \ref{thm:rdp_vector}.

\begin{theorem}
\label{thm:rdp_vector}
Let $B_\text{in}=2$ and let $M$ be the Fisher information constant from Lemma \ref{lem:rdp_scalar}. Suppose that $\bx_1, \bx_2 \in \mathbb{R}^d$ satisfy $\| \bx_2 - \bx_1 \|_2 \leq C$. Then the I-MVU mechanism is $(\alpha, \alpha M C^2 / 2)$-RDP for all $\alpha > 1$.
\end{theorem}

\textbf{Proof sketch.} We first derive the Taylor series expression for the R\'{e}nyi divergence between $\mathbb{P}(\cdot | \bseta(x_1))$ and $\mathbb{P}(\cdot | \bseta(x_2))$. Since R\'{e}nyi divergence is minimized and is equal to $0$ when $x_1 = x_2$, the zeroth-order and first-order terms in the Taylor series are $0$. The coefficient for the second-order term is given by the Fisher information $\calI_Z(x_1)$~\cite{haussler1997mutual}, and thus we give a numerical method to compute $M = \sup_{x \in \mathbb{R}} \calI_Z(x)$ in Appendix \ref{sec:computing_fil} and use it in \autoref{eq:rdp_bound} to bound the RDP $\epsilon$.

%% file: experiment.tex
\section{Experiments}

\begin{figure*}[t]
    \centering
    \begin{subfigure}{.49\textwidth}
      \centering
      \includegraphics[width=\linewidth]{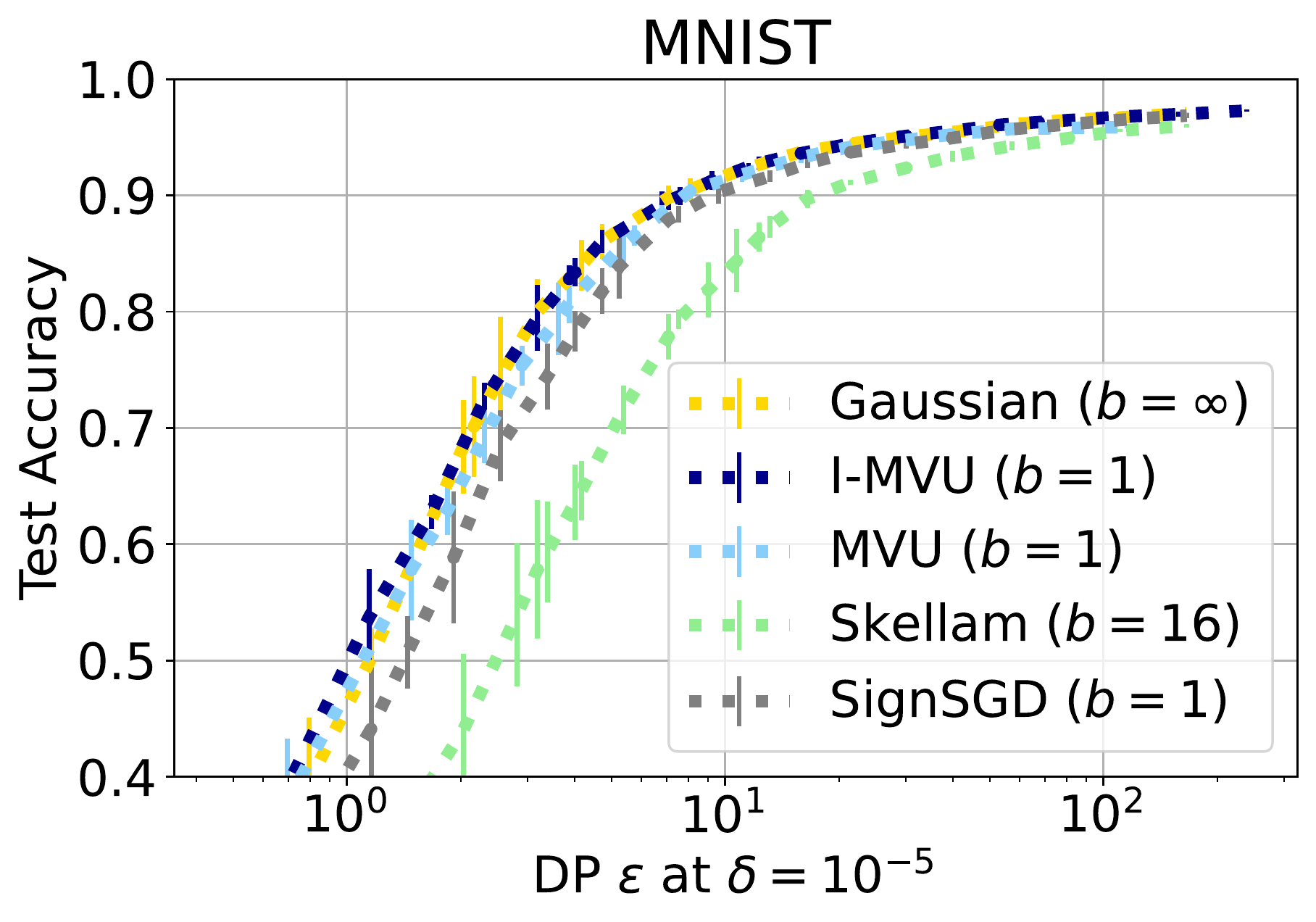}  
    \end{subfigure}
    \begin{subfigure}{.49\textwidth}
      \centering
      \includegraphics[width=\linewidth]{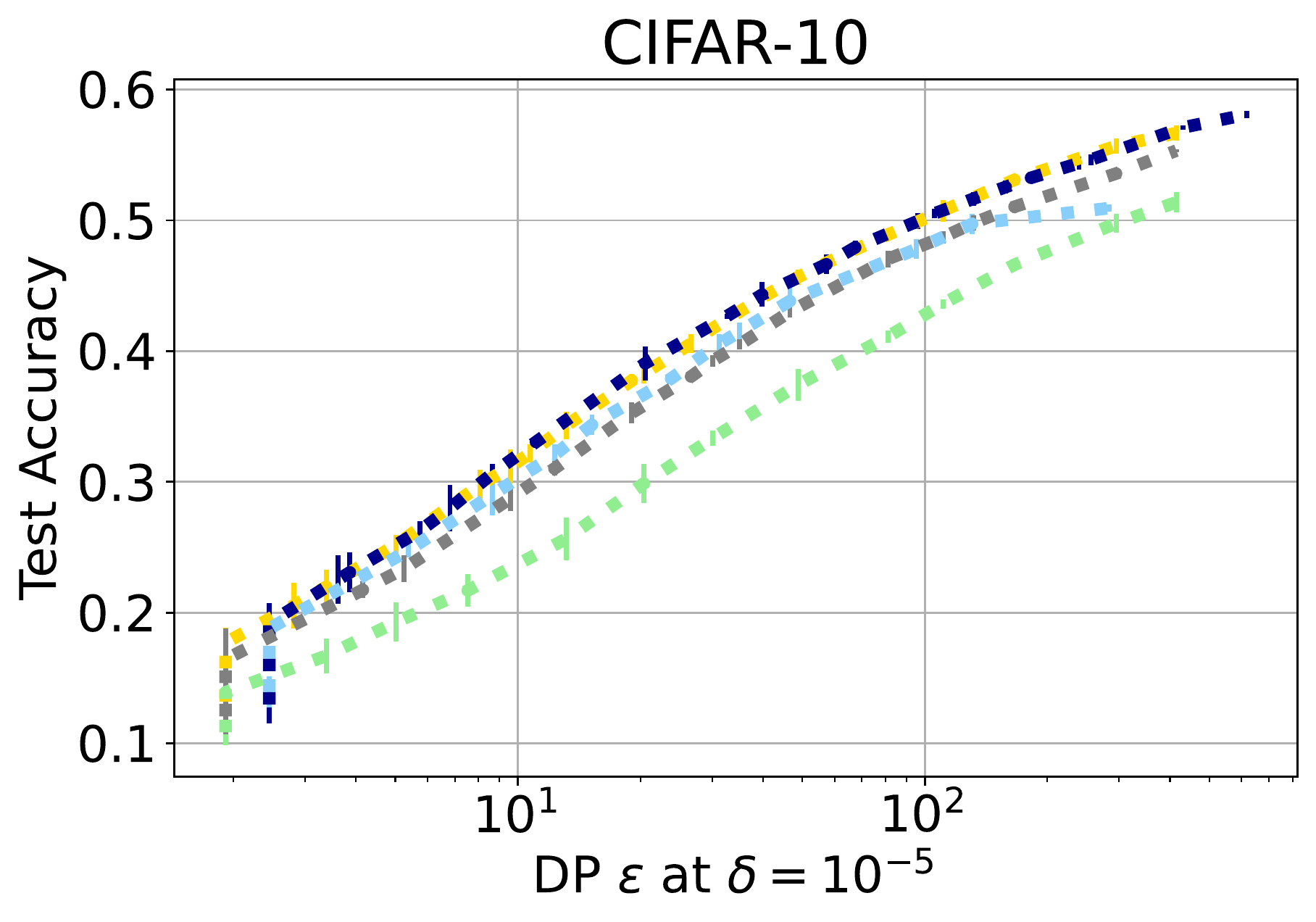}  
      \end{subfigure} 
\vspace{-1ex}
\caption{Privacy vs. accuracy plot for the client-level DP setting on MNIST (left) and CIFAR-10 (right) for $L_2$-bounded client updates. Each line shows the best privacy-utility curve across a grid of hyperparameters, and vertical bar shows standard deviation. I-MVU consistently outperforms all other baselines and achieves the same privacy-utility trade-off as the non-compressed Gaussian mechanism while requiring only one bit communication per parameter.}
\label{fig:dpsgd_comparison}
\vspace{-1ex}
\end{figure*}

\begin{figure}[t]
    \centering
    \includegraphics[width=\linewidth]{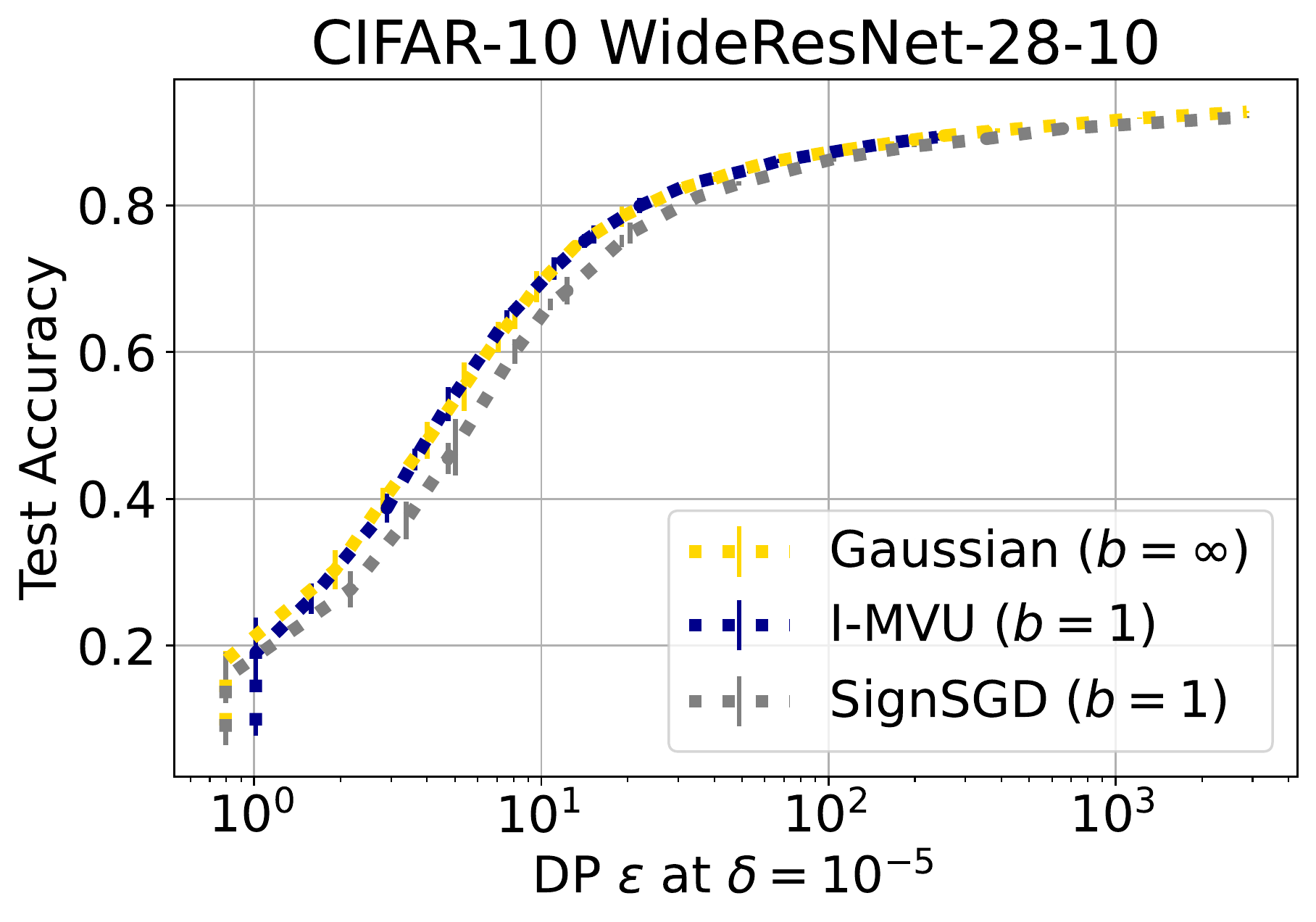}
\vspace{-3ex}
\caption{Privacy vs. accuracy plot for the client-level DP setting on CIFAR finetuning with an ImageNet-pretrained WideResNet-28-10 model. Each line shows the best privacy-utility curve across a grid of hyperparameters, and vertical bar shows standard deviation.}
\label{fig:dpsgd_cifar_wrn}
\end{figure}

We evaluate the I-MVU mechanism for federated learning under the local DP setting, \emph{i.e.}, clients transmit the privately compressed model update $\calM(\bx)$ to the server \emph{before} aggregation. We consider two different notions of adjacency for DP: \emph{client-level}, where adjacency is defined as removal of any client, and \emph{sample-level}, where adjacency is defined as removal of a single sample at one client.

\paragraph{Baselines.} We compare I-MVU with several privacy-aware compression mechanisms that have been proposed for private FL training:
\begin{enumerate}
\item \emph{MVU mechanism}~\citep{chaudhuri2022privacy} for $L_2$-bounded client updates.
\item \emph{Skellam mechanism}~\citep{agarwal2021skellam}, which discretizes the input using randomized dithering and adds noise from the Skellam distribution to achieve privacy-aware compression of $L_2$-bounded client updates.
\item \emph{SignSGD}~\citep{jin2020stochastic} is a strong simple baseline that first applies a non-compressed DP mechanism to the gradient and then transmits its sign coordinate-wise. Since it operates as a post-processing step on top of a non-compressed DP mechanism, it inherits the privacy guarantee of the non-compressed mechanism while only outputting one bit per parameter. We evaluate SignSGD with both Laplace and Gaussian mechanisms for $L_1$-bounded and $L_2$-bounded client updates.
\item \emph{Laplace and Gaussian mechanisms} serve as the non-compressed baselines for $L_1$-bounded and $L_2$-bounded client updates. We transmit the private updates using floating point with 32 bits per parameter.
\end{enumerate}
A good privacy-aware compression mechanism should achieve a desirable privacy-utility trade-off (\emph{i.e.}, high test accuracy with low DP $\epsilon$) while transmitting the client updates in a communication-efficient manner. 
Since our application domain is FL, we did not consider many generic DP distributed mean estimation mechanisms such as PrivUnit~\citep{bhowmick2018protection} and Poisson binomial mechanism~\citep{chen2022poisson}. These mechanisms are designed primarily for privately compressing low-dimensional vectors in an unbiased manner; in contrast, model updates in FL are high-dimensional $L_2$-bounded vectors and do not need to be unbiased.

\begin{figure*}[t]
    \centering
    \begin{subfigure}{.49\textwidth}
      \centering
      \includegraphics[width=\linewidth]{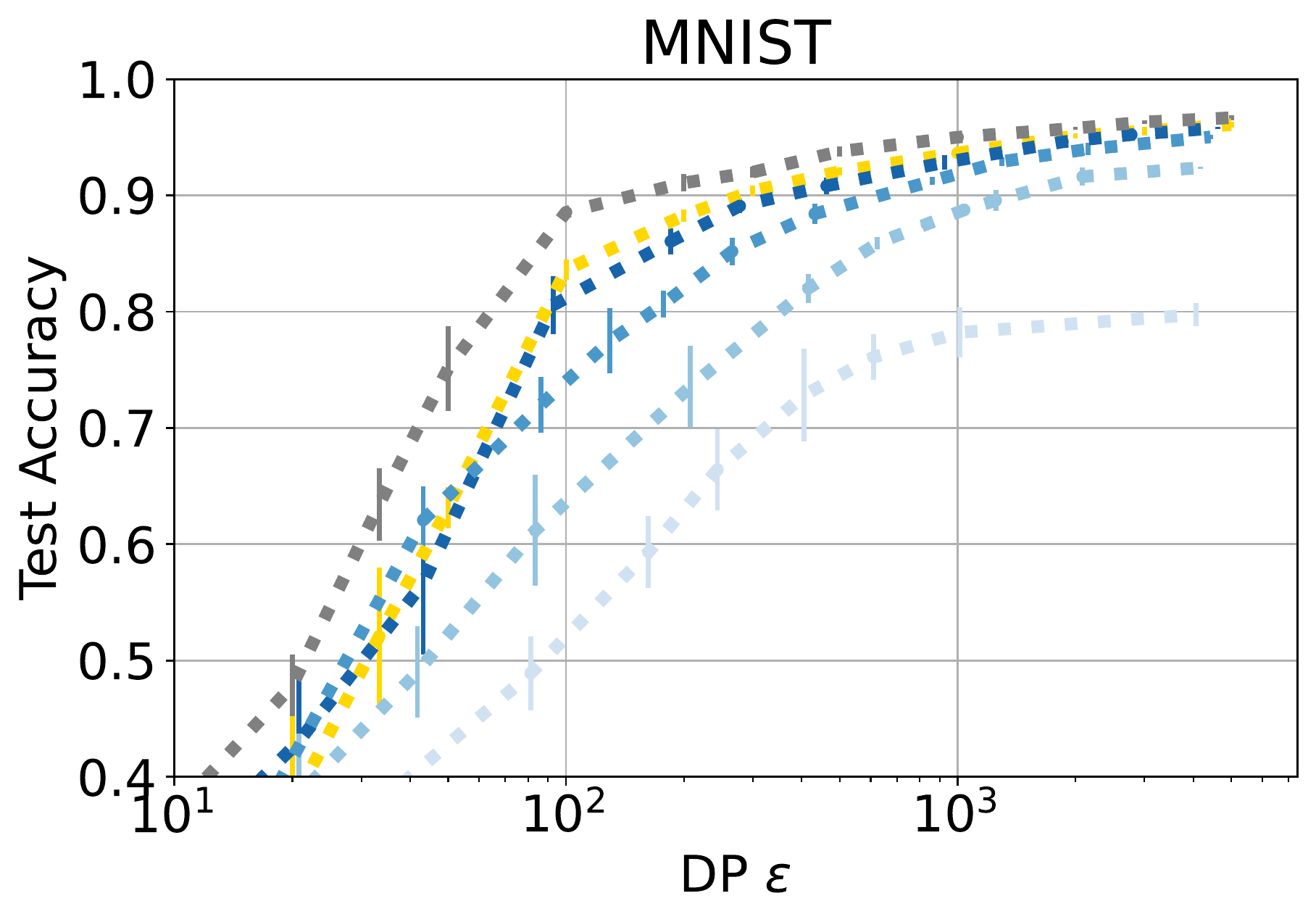}  
    \end{subfigure}
    \begin{subfigure}{.49\textwidth}
      \centering
      \includegraphics[width=\linewidth]{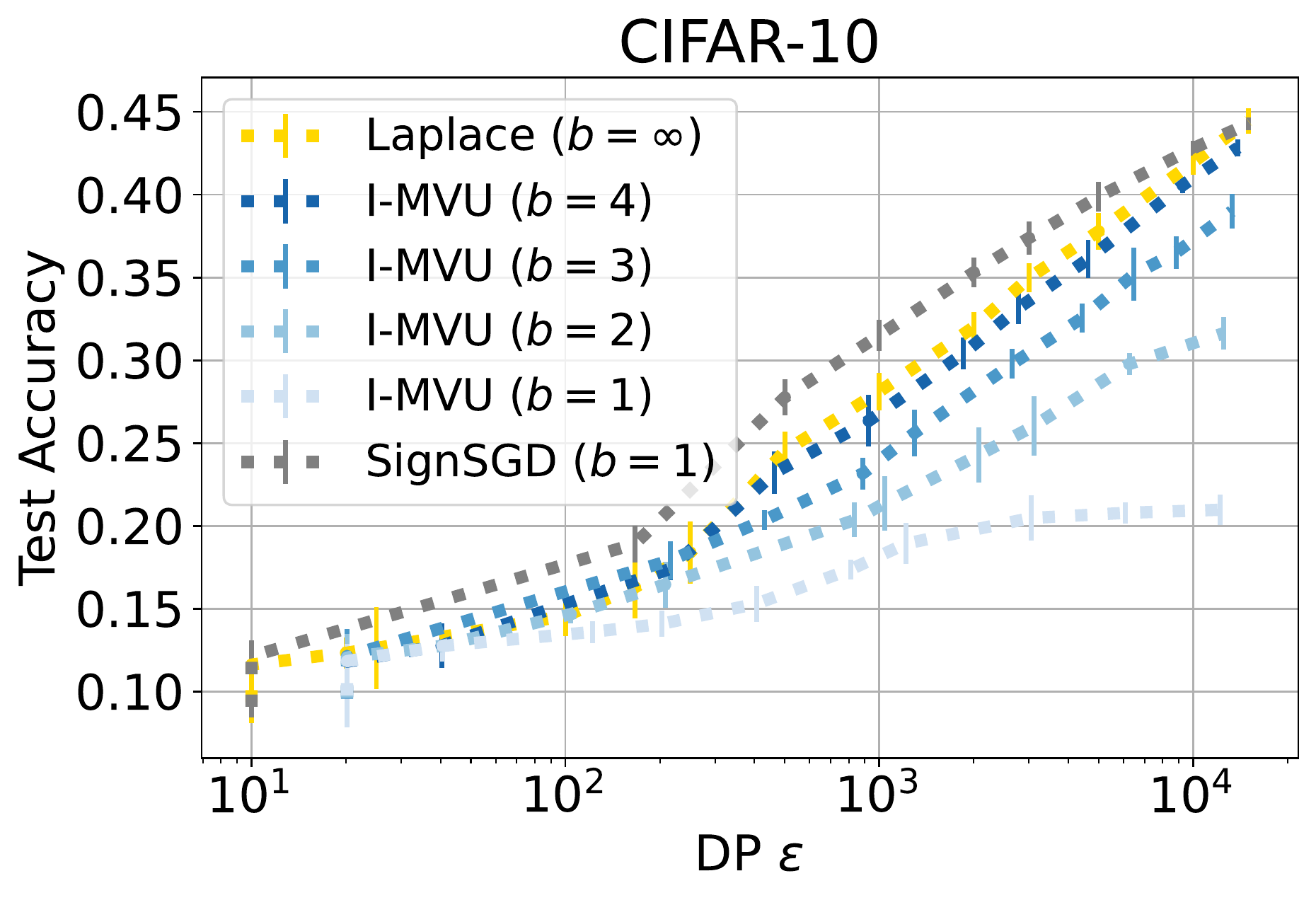}  
      \end{subfigure} 
\vspace{-1ex}
\caption{Privacy vs. accuracy plot for the client-level DP setting on MNIST (left) and CIFAR-10 (right) for $L_1$-bounded client updates. Each line shows the best privacy-utility curve across a grid of hyperparameters, and vertical bar shows standard deviation. I-MVU with $b=4$ achieves the same privacy-utility trade-off as the non-compressed Laplace mechanism, while decreasing the communication budget $b$ worsens the trade-off. Surprisingly, SignSGD consistently outperforms the non-compressed Laplace baseline, possibly due to the variance reduction (and bias-introducing) effect of sign compression.}
\label{fig:dpsgd_l1_comparison}
\vspace{-1ex}
\end{figure*}

\subsection{Client-level DP}
\label{sec:client_dp}

We first evaluate under the \emph{client-level DP} setting on MNIST and CIFAR-10~\cite{krizhevsky2009learning}. Here, the privacy analysis guarantees that the learning algorithm is differentially private with respect to the removal of any client. We divide the training set among the clients with client sample size 1. Each client performs a single local gradient update in every FL round. This setting is equivalent to DP-SGD training~\cite{abadi2016deep} but with the Gaussian mechanism replaced by a communication-efficient private mechanism.

\textbf{Training details.} Following \cite{chaudhuri2022privacy}, we train a linear model on top of ScatterNet features~\cite{tramer2020differentially}. This training recipe remains highly competitive under the central DP setting for MNIST and CIFAR-10 without leveraging any public data, hence we adopt it for FL training under local DP.
Following \citet{abadi2016deep}, we apply $L_1$ and $L_2$ gradient norm clipping to control the gradient sensitivity and then apply a privacy-aware compression mechanism to transmit the clipped gradient privately.
We perform a grid search over hyperparameters such as number of update rounds, step size, gradient norm clip, and mechanism parameters $\sigma$ (for Gaussian and SignSGD) and $\epsilon$ (for MVU and I-MVU); see \autoref{sec:hyerparameters} for details. 

\textbf{Result for $L_2$-bounded client updates.} We first consider the more common approach of $L_2$-norm clipping. Figure \ref{fig:dpsgd_comparison} shows the privacy vs.~test accuracy curve on MNIST (left) and CIFAR-10 (right). Privacy is measured in terms of $(\epsilon,\delta)$-DP at $\delta=10^{-5}$. 
For each hyperparameter setting, we compute the average test accuracy over five training runs to produce a single $\epsilon$-accuracy pair, and plot the Pareto frontier as a dotted line.
The yellow line corresponds to the standard Gaussian mechanism without compression, which attains the best test accuracy at any given privacy budget $\epsilon$. Both SignSGD and MVU are competitive, achieving close to the same level of accuracy as the Gaussian mechanism, but a non-negligible gap remains, especially on CIFAR-10. In contrast, I-MVU attains nearly the same performance as the Gaussian mechanism at all values of $\epsilon$ on both MNIST and CIFAR-10. Since MVU and I-MVU are near-identical mechanisms, we argue that the performance gain comes primarily from the tight privacy analysis for $L_2$ geometry using Fisher information (Section \ref{sec:analysis}).

In Figure \ref{fig:dpsgd_cifar_wrn} we repeat the experiment with an ImageNet pre-trained WideResNet-28-10~\citep{zagoruyko2016wide} model. The model is pre-trained on down-scaled $32 \times 32$ resolution ImageNet~\citep{deng2009imagenet} samples and then finetuned on CIFAR-10 using either the Gaussian mechanism, signSGD or I-MVU. We did not apply the MVU or Skellam mechanisms because the dithering operation is prohibitively expensive on such a large model. We observe a similar result where I-MVU matches the performance of the Gaussian mechanism across the board. 

\textbf{Result for $L_1$-bounded client updates.} For completeness, we also show the result for compressing updates with $L_1$-norm clipping. This approach is uncommon in private FL training as the noise variance is much larger, but can provide an $\epsilon$-DP guarantee as opposed to $(\epsilon,\delta)$-DP with $L_2$-norm clipping. Figure \ref{fig:dpsgd_l1_comparison} shows the privacy (in terms of $\epsilon$-DP) vs.~test accuracy Pareto frontier curve on MNIST (left) and CIFAR-10 (right). It can be seen that I-MVU with $b=4$ attains almost the same privacy-utility trade-off as the non-compressed Laplace mechanism, while I-MVU with lower $b$ attain a worse trade-off. Surprisingly, SignSGD outperforms both Laplace and I-MVU with a communication budget of $b=1$. We suspect this is due to the inherent variance reduction effect of sign compression and that FL training does not require the gradient to be unbiased.

\subsection{Sample-level DP}
\label{sec:sample_dp}

\begin{figure}[t]
    \centering
    \includegraphics[width=\linewidth]{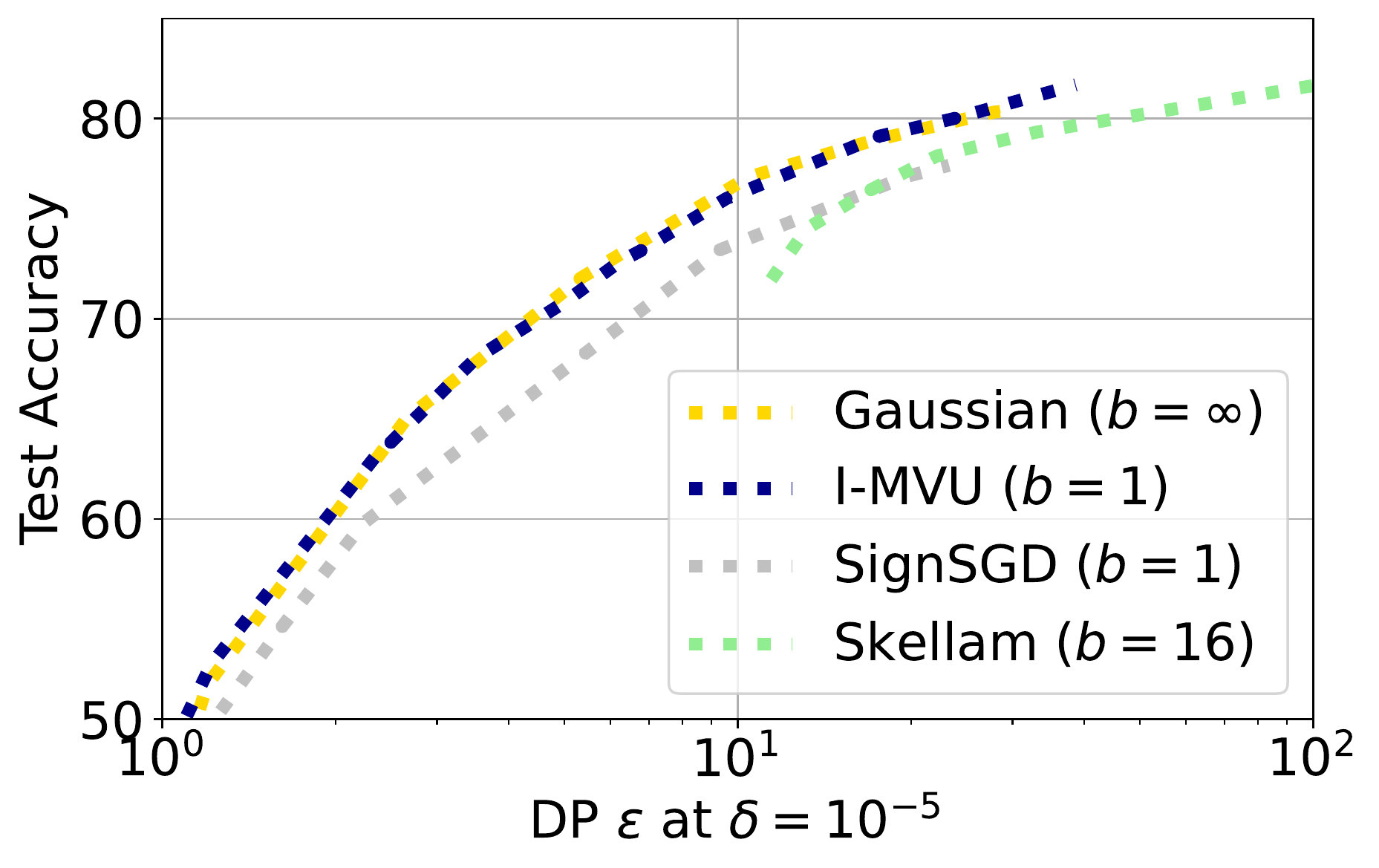}
\vspace{-3ex}
\caption{Privacy vs. accuracy plot for the sample-level DP scenario on FEMNIST. I-MVU with one-bit communication budget per coordinate consistently performs better than SignSGD and is competitive with the non-compressed Gaussian baseline across the entire range of $\epsilon$. Finally, I-MVU outperforms the Skellam mechanism while using 16 times less communication bandwidth.}
\label{fig:dpsgd_femnist}
\vspace{-1ex}
\end{figure}

Next, we evaluate under the \emph{sample-level DP} setting on the FEMNIST dataset~\cite{caldas2018leaf} for classifying written characters into 62 distinct classes. Privacy analysis guarantees that the learning algorithm is differentially private with respect to the removal of \emph{any training sample from a client}. The dataset has a pre-defined train split with $3,500$ clients, from which we randomly select $3,150$ clients for training and the remaining $350$ clients for testing. A set of $5$ clients is selected in each training round. Each client performs full batch gradient descent for a single local gradient update to compute the update vector. The update vector is privatized using a communication-efficient private mechanism and transmitted to the server.

\textbf{Training details.} We train a simple 4-layer convolutional network with 46k parameters for classification. The model achieves $84\%$ accuracy when trained non-privately with FL. The client optimizer is SGD with a learning rate of $0.1$ and no momentum. The server implements FedAvg~\cite{mcmahan2017communication} with a momentum of $0.9$. We perform a grid search on the server learning rate, the clipping factor, the noise multiplier $\sigma$ for Gaussian and SignSGD, noise scale $\mu$ for Skellam, and the $L_1$-metric DP $\epsilon$ and scale hyperparameters for I-MVU. 
The hyperparameter ranges are given in Tables~\ref{tab:femnist_hyperparams_imvu}, \ref{tab:femnist_hyperparams_gaussian} and \ref{tab:femnist_hyperparams_ssgd} in the appendix. In particular, SignSGD uses a much lower server-side learning rates since the updates (in $\{\pm 1\}$) have higher magnitude.

\textbf{Result.} We show the privacy-accuracy trade-off for FEMNIST in Figure~\ref{fig:dpsgd_femnist}. The Pareto frontier represents the optimal privacy-accuracy trade-off. The DP privacy budget $\epsilon$ is given at $\delta = 10^{-5}$. We observe that I-MVU (blue dashed line) performs better than SignSGD (silver dashed line) for the same communication budget of one bit per update coordinate across the entire range of considered privacy budgets $\epsilon$. Moreover, I-MVU performs on par with the non-compressed Gaussian baseline (yellow dashed line), where clients perform local DP-SGD without compressing model updates. I-MVU also outperforms the Skellam mechanism (green dashed line) using 16 times less bandwidth. Additionally, to demonstrate the performance of I-MVU across a wider bit range, we display the performance of the proposed method for various communication budgets in Figure~\ref{fig:dpsgd_multiple_bits}.

%% file: relatedwork.tex
\section{Related Work}
\label{sec:related}

\paragraph{Privacy-aware compression.} The problem of privacy-aware compression for distributed mean estimation (DME) has received much attention in recent years.
Prior work proposed notable mechanisms such as PrivUnit~\cite{bhowmick2018protection, asi2022optimal}, discrete Gaussian~\cite{canonne2020discrete, kairouz2021distributed}, Skellam~\cite{agarwal2021skellam}, MVU~\cite{chaudhuri2022privacy} and others~\cite{chen2020breaking, girgis2021shuffled, shah2022optimal, chen2022poisson, lang2022joint} for computing the mean of a set of vectors under communication constraint.  \citet{feldman2021lossless} provides a general procedure for converting a non-compressed mechanism to a compressed one that is computationally differentially private under the assumption that the pseudo-random generator in the privacy mechanism is hard to break. 


There are also other approaches for privacy-aware compression in FL that do not rely on solving the DME problem.
For example, \citet{jin2020stochastic} first showed that a simple post-processing of non-compressed mechanisms by taking the sign is a strong baseline that does not attain unbiasedness but works well in practice. Other mechanisms such as DP-REC~\cite{triastcyn2021dp} have also leveraged special properties of FL to design more efficient privacy-aware compression schemes for transmitting model updates.



\paragraph{Non-private compression in FL.}
Previous work has also studied non-private compression in FL, with a focus on characterizing how noise due to (lossy) compression impacts optimization dynamics. Methods have been proposed to only compress up-link (from client to server) messages~\cite{basu2019qsparse,reisizadeh2020fedpaq,albasyoni2020optimal,haddadpour2021federated,mitra2021linear,sadiev2022federated}, as well as both up- and down-link messages~\cite{philippenko2020artemis,condat2022provably}. Although none of these provides privacy, they do illustrate that federated training is possible with biased compression methods. Note that, in private FL, the downlink messages can be considered non-private, so only the uplink would need to make use of privacy-aware compression.

\paragraph{Mechanism design through optimization.} Classical DP mechanisms are mostly derived from simple parameterized distributions and whose privacy is analyzed mathematically. For more specialized applications such as FL where the requirements are complex, it is often helpful to design custom mechanisms by optimizing a certain objective under DP constraints. \citet{geng2015staircase} proposed the staircase mechanism and showed that it is variance-optimal in the low-to-medium privacy regime for privatizing $L_1$-bounded vectors. \citet{bhowmick2018protection, asi2022optimal} proposed the PrivUnit mechanism and its variant PrivUnitG that are variance-optimal for privatizing unit $L_2$ vectors under a communication constraint. More similar to our work, \citet{bordenabe2014optimal} designed an optimal mechanism for location privacy by numerically solving a linear program.

%% file: conclusion.tex
\section{Conclusion}
\label{sec:conclusion}

We proposed the Interpolated MVU (I-MVU) mechanism that drastically reduces the amount of uplink communication in cross-device FL while providing differential privacy guarantees. I-MVU can be readily applied to DP training of FL models with $L_1$- and $L_2$-bounded vectors, with performing matching that of the Laplace and Gaussian mechanisms while communicating as few as one bit per parameter.

\paragraph{Limitations.} We discuss several limitations of I-MVU and suggest potential directions for future work.

1. The mechanism currently operates under the local DP setting, which does not leverage the power of secure aggregation to further reduce the privacy cost. Since I-MVU outputs samples from the categorical distribution, their sum forms a so-called Poisson multinomial distribution, whose properties need to be analyzed to understand the privacy cost of I-MVU under the central DP setting.

2. Our work focused on extending MVU through better interpolation and privacy analysis and did not modify the underlying mechanism design problem. Choices such as the set of grid points $\mathcal{X}$, variance objective, and DP constraint can be adapted to better leverage the structure of the new interpolation scheme.

3. Our scalar-to-vector extension of I-MVU can achieve at best a one bit per parameter compression rate. Compressing further to below one bit per parameter may be necessary for applications of FL to large-scale models. Combining I-MVU with recent work on sketching-based compression~\citep{chen2022fundamental} can be a promising solution.

%% file: appendix.tex
\section{Proofs}
\label{sec:proofs}

\begin{lemma}
Suppose $x, x' \in [0, 1]$ and the MVU mechanism satisfies $\epsilon$ $L_1$-metric-DP. Let $\epsilon' = (B_\text{in} - 1) \cdot (\max_i \max_{\xi \in [x_i, x_{i+1}]} |\sigma(\bseta(\xi))^\top (\bseta_{i+1} - \bseta_i)|)$. Then:
\begin{equation*}
    D_\infty(\mathbb{P}(\cdot|\bseta(x)) ~||~ \mathbb{P}(\cdot|\bseta(x'))) \leq (\epsilon + \epsilon') |x' - x|.
\end{equation*}
\end{lemma}
\begin{proof}
Suppose first that $x, x' \in [x_i, x_{i+1}]$ for some $i$. Recall that
\begin{equation*}
    \bseta(x) = \left( \frac{x_{i+1} - x}{\Delta} \right) \bseta_i + \left( \frac{x - x_i}{\Delta} \right) \bseta_{i+1}.
\end{equation*}
Using \autoref{eq:natural_param}, we get that for any $j$:
\begin{align}
    | \log \mathbb{P}(j|\bseta(x)) - \log \mathbb{P}(j|\bseta(x')) | &= \left| \left( \frac{x' - x}{\Delta} \right) \be_j^\top \bseta_i + \left( \frac{x - x'}{\Delta} \right) \be_j^\top \bseta_{i+1} - A(\bseta(x)) + A(\bseta(x')) \right| \nonumber \\
    &= \left| \left( \frac{x' - x}{\Delta} \right) \be_j^\top (\bseta_i - \bseta_{i+1}) - A(\bseta(x)) + A(\bseta(x')) \right|. \label{eq:max_log_diff}
\end{align}
Since $A'(\bseta(x)) = \sigma(\bseta(x))$, we can apply Taylor expansion to $A$ to get:
\begin{equation*}
    A(\bseta(x')) = A(\bseta(x)) + \left( \frac{x' - x}{\Delta} \right) \sigma(\bseta(\xi))^\top (\bseta_{i+1} - \bseta_i),
\end{equation*}
where $\xi \in [x_i, x_{i+1}]$. Substituting into \autoref{eq:max_log_diff} gives:
\begin{align*}
    | \log \mathbb{P}(j|\bseta(x)) - \log \mathbb{P}(j|\bseta(x')) | &= \left| \left( \frac{x' - x}{\Delta} \right) \be_j^\top (\bseta_i - \bseta_{i+1}) + \left( \frac{x' - x}{\Delta} \right) \sigma(\bseta(\xi))^\top (\bseta_{i+1} - \bseta_i) \right| \\
    &= \left| \left( \frac{x' - x}{\Delta} \right) (\sigma(\bseta(\xi)) - \be_j)^\top (\bseta_{i+1} - \bseta_i) \right| \\
    &\leq (B_\text{in} - 1) \cdot \left( |\sigma(\bseta(\xi))^\top (\bseta_{i+1} - \bseta_i)| + | \be_j^\top (\bseta_{i+1} - \bseta_i) | \right) | x' - x |.
\end{align*}
Since the MVU mechanism is $\epsilon$ $L_1$-metric-DP, we have
$|\be_j^\top (\bseta_{i+1} - \bseta_i)| = | \bseta_{i+1,j} - \bseta_{i,j} | \leq \epsilon |x_{i+1} - x_i| = \epsilon / (B_\text{in} - 1)$. Thus:
$$| \log \mathbb{P}(j|\bseta(x)) - \log \mathbb{P}(j|\bseta(x')) | \leq \left( (B_\text{in}-1) \cdot |\sigma(\bseta(\xi))^\top (\bseta_{i+1} - \bseta_i)| + \epsilon \right) | x' - x | \leq (\epsilon + \epsilon') | x' - x |,$$ as desired. Suppose now that $x \in [x_i, x_{i+1}]$ and $x' \in [x_k, x_{k+1}]$ with $i \leq k$. We can write:
\begin{align*}
\log \mathbb{P}(j|\bseta(x)) - \log \mathbb{P}(j|\bseta(x')) &\leq |\log \mathbb{P}(j|\bseta(x)) - \log \mathbb{P}(j|\bseta(x_{i+1}))| \\
&+ \sum_{l=i+1}^{k-1} |\log \mathbb{P}(j|\bseta(x_l)) - \log \mathbb{P}(j|\bseta(x_{l+1}))| + |\log \mathbb{P}(j|\bseta(x_k)) - \log \mathbb{P}(j|\bseta(x'))| \\
&\leq (\epsilon + \epsilon') \left( |x - x_{i+1}| + \sum_{l=i+1}^{k-1} | x_l - x_{l+1} | + |x_k - x'| \right) \\
&= (\epsilon + \epsilon') |x - x'|.
\end{align*}
A similar argument applies when $i > k$.
\end{proof}

\begin{theorem}
Let $\epsilon'$ be the constant from Lemma \ref{lem:dp_scalar}. Suppose the MVU mechanism is $\epsilon$ $L_1$-metric-DP and that $\bx, \bx' \in \mathbb{R}^d$ satisfy $\| \bx' - \bx \|_1 \leq C$, then the I-MVU mechanism is $(\epsilon + \epsilon') C$-DP.
\end{theorem}
\begin{proof}
Let $\ba = \calM(\bx) \in \{a_1,\ldots,a_{B_\text{out}}\}^d$ be the output of the vector I-MVU mechanism that independently applies the scalar mechanism to each coordinate. Then:
\begin{align*}
    D_\infty(\calM(\bx) ~||~ \calM(\bx')) &= \max_{\ba \in \{a_1,\ldots,a_{B_\text{out}}\}^d} \left| \sum_{k=1}^d \log \mathbb{P}(\ba_k | \bseta(\bx'_k)) - \sum_{k=1}^d \log \mathbb{P}(\ba_k | \bseta(\bx_k)) \right| \\
    &\leq \sum_{k=1}^d \max_{\ba_k \in \{a_1,\ldots,a_{B_\text{out}}\}} \left| \log \mathbb{P}(\ba_k | \bseta(\bx'_k)) - \log \mathbb{P}(\ba_k | \bseta(\bx_k)) \right| \\
    &= \sum_{k=1}^d D_\infty(\mathbb{P}(\cdot | \bseta(\bx_k)) ~||~ \mathbb{P}(\cdot | \bseta(\bx'_k))) \\
    &\leq \sum_{k=1}^d \tilde{\Delta}(\bseta) |\bx'_k - \bx_k| \quad \text{by Lemma \ref{lem:dp_scalar}}\\
    &= \tilde{\Delta}(\bseta) \| \bx' - \bx \|_1.
\end{align*}
\end{proof}

\begin{lemma}
Let $B_\text{in}=2$ and let $M = \sup_{x \in \mathbb{R}} \calI_Z(x)$ be an upper bound on the Fisher information of the mechanism $\calM$. Then for any $x, x' \in \mathbb{R}$:
\begin{equation*}
    D_\alpha(\mathbb{P}(\cdot|\bseta(x)) ~||~ \mathbb{P}(\cdot|\bseta(x'))) \leq \alpha M (x' - x)^2 / 2.
\end{equation*}
\end{lemma}
\begin{proof}
We first derive an explicit form for the Fisher information. Let $f(\bz; x)$ denote the pmf in \autoref{eq:natural_param} for any $\bz \in \{\be_1,\ldots,\be_{B_\text{out}}\}$. The log pmf is:
\begin{equation}
    \log f(\bz; x) = \bz^\top \bseta(x) - A(\bseta(x))
\end{equation}
 Taking derivative with respect to $x$ gives:
\begin{align*}
    \frac{d}{dx} \log f(\bz; x) &= (\bz - \sigma(\bseta(x)))^\top (\bseta_2 - \bseta_1), \label{eq:log_density_grad} \\
    \left( \frac{d}{dx} \log f(\bz; x) \right)^2 &= (\bseta_2 - \bseta_1)^\top (\bz - \sigma(\bseta(x))) (\bz - \sigma(\bseta(x)))^\top (\bseta_2 - \bseta_1),
\end{align*}
where $\sigma$ denotes the sigmoid function. Taking expectation over $\bz$ gives the Fisher information:
\begin{equation}
    \label{eq:fisher_information}
    \mathcal{I}_Z(x) = (\bseta_2 - \bseta_1)^\top U (\bseta_2 - \bseta_1),
\end{equation}
with $U = \text{diag}(\sigma(\bseta(x))) - \sigma(\bseta(x)) \sigma(\bseta(x))^\top$.

To derive the upper bound, we first define a function $F_\alpha$ for the R\'{e}nyi divergence of the mechanism for a fixed $x$ and varying $x'$:
\begin{equation}
    \label{eq:renyi_function}
    F_\alpha(x';x) = D_\alpha(\mathbb{P}(\cdot|\bseta(x)) ~||~ \mathbb{P}(\cdot|\bseta(x'))).
\end{equation}
By Taylor's theorem, we can express $F_\alpha$ as:
\begin{equation*}
    F_\alpha(x';x) = F_\alpha(x;x) + (x' - x) F'_\alpha(x;x) + (x' - x)^2 F''_\alpha(\xi;x) / 2
\end{equation*}
for some $\xi \in [x,x']$. Note that $F_\alpha(x;x) = 0$ and $F'_\alpha(x;x) = 0$ (since $x$ is the global minimizer of $F_\alpha(\cdot;x)$), so $F_\alpha$ is locally a quadratic function:
\begin{equation}
    \label{eq:quadratic_func}
    F_\alpha(x';x) = (x' - x)^2 F''_\alpha(\xi;x) / 2.
\end{equation}
Since $f$ is the pmf of an exponential family distribution, we can use the closed form expression~\cite{nielsen2011r} for R\'{e}nyi divergence of exponential family distributions to express $F_\alpha$ and its derivatives:
\begin{align*}
    F_\alpha(\xi; x) &= \frac{1}{\alpha-1}[A(\alpha \bseta(x) + (1-\alpha) \bseta(\xi)) - \alpha A(\bseta(x)) - (1-\alpha) A(\bseta(\xi))] \\
    F_\alpha'(\xi; x) &= (\sigma(\bseta(\xi)) - \sigma(\alpha \bseta(x) + (1-\alpha) \bseta(\xi)))^\top (\bseta_2 - \bseta_1) \\
    F_\alpha''(\xi; x) &= (\bseta_2 - \bseta_1)^\top (V + (\alpha-1) W) (\bseta_2 - \bseta_1)
\end{align*}
where $V = \text{diag}(\sigma(\bseta(\xi))) - \sigma(\bseta(\xi)) \sigma(\bseta(\xi))^\top$, $W = \text{diag}(\sigma(\bseta(x''))) - \sigma(\bseta(x'')) \sigma(\bseta(x''))^\top$, and $x'' = \alpha x + (1-\alpha) \xi$. Hence $F_\alpha''(\xi; x) = \mathcal{I}_Z(\xi) + (\alpha-1) \mathcal{I}_Z(x'')$ by \autoref{eq:fisher_information}. Upper bounding $\calI_Z(\xi)$ and $\calI_Z(x'')$ by $M := \sup_{x \in \mathbb{R}} \calI_Z(x)$ and combining with \autoref{eq:quadratic_func} gives the desired result.
\end{proof}

\begin{theorem}
Let $B_\text{in}=2$ and let $M$ be the Fisher information constant from Lemma \ref{lem:rdp_scalar}. Suppose that $\bx, \bx' \in \mathbb{R}^d$ satisfy $\| \bx' - \bx \|_2 \leq C$, then the I-MVU mechanism is $(\alpha, \alpha M C^2 / 2)$-RDP for all $\alpha > 1$.
\end{theorem}
\begin{proof}
Let $\ba = \calM(\bx) \in \{a_1,\ldots,a_{B_\text{out}}\}^d$ be the output of the vector I-MVU mechanism that independently applies the scalar mechanism to each coordinate. Then:
\begin{align*}
    D_\alpha(\calM(\bx) ~||~ \calM(\bx')) &= \frac{1}{\alpha-1} \log \sum_{\ba \in \{a_1,\ldots,a_{B_\text{out}}\}^d} \prod_{k=1}^d \frac{\mathbb{P}(\ba_k | \bseta(\bx'_k))^\alpha}{\mathbb{P}(\ba_k | \bseta(\bx_k))^{\alpha-1}} \\
    &= \sum_{k=1}^d \frac{1}{\alpha-1} \log \sum_{\ba_k \in \{a_1,\ldots,a_{B_\text{out}}\}} \frac{\mathbb{P}(\ba_k | \bseta(\bx'_k))^\alpha}{\mathbb{P}(\ba_k | \bseta(\bx_k))^{\alpha-1}} \quad \text{by independence}\\
    &= \sum_{k=1}^d D_\alpha(\mathbb{P}(\cdot | \bseta(\bx_k)) ~||~ \mathbb{P}(\cdot | \bseta(\bx'_k))) \\
    &\leq \sum_{k=1}^d \alpha M (\bx'_k - \bx_k)^2 / 2 \quad \text{by Lemma \ref{lem:rdp_scalar}}\\
    &= \alpha M \| \bx' - \bx \|_2^2 / 2.
\end{align*}
\end{proof}

\section{Computing Fisher Information}
\label{sec:computing_fil}

In this section we describe a method for computing $M = \sup_{x \in \mathbb{R}} \calI_Z(x)$. We first define a condition for $\bseta_1, \bseta_2$ that allows us to reduce this problem to maximizing $\calI_Z(x)$ over a bounded range $[x_{\min}, x_{\max}]$.

\begin{definition}
\label{def:anadrome}
Two vectors $\bseta_1, \bseta_2 \in \mathbb{R}^B$ are said to be \emph{anadromic} if for all $j=1,\ldots,B$, we have $(\bseta_1)_j = (\bseta_2)_{B-j+1}$.
\end{definition}

The following technical lemma proves several useful properties that hold when $\bseta_1$ and $\bseta_2$ are anadromic.

\begin{lemma}
\label{lem:fisher_anadromic}
Suppose that $\bseta_1, \bseta_2 \in \mathbb{R}^B$ are anadromic. Let $\bstheta = \bseta_2 - \bseta_1$ and suppose that $j^+ = \argmax_j \bstheta_j, j^- = \argmin_j \bstheta_j$ are unique. Then the following hold:
\begin{enumerate}[label=(\roman*), noitemsep, leftmargin=*, topsep=0pt]
\item $\bstheta_j = - \bstheta_{B-j+1}$ for all $j$, and hence $j^- = B- j^+ + 1$.
\item $\bseta(x)_j = \bseta(1-x)_{B-j+1}$ for all $j$.
\item $\sigma(\bseta(x)) \rightarrow \be_{j^+}$ as $x \rightarrow \infty$ and $\sigma(\bseta(x)) \rightarrow \be_{j^-}$ as $x \rightarrow -\infty$.
\item $\mathcal{I}_Z(x) = \mathcal{I}_Z(1-x)$ for all $x \in \mathbb{R}$.
\item $x = 1/2$ is a stationary point for $\mathcal{I}_Z(x)$.
\item If $\sigma(\bseta(x))_{j^+} \geq 1/2$ then $\calI_Z(x) \leq 4 \bstheta_{j^+}^2 \sigma(\bseta(x))_{j^+} (1 - \sigma(\bseta(x))_{j^+})$.
\end{enumerate}
\end{lemma}
\begin{proof}
(i) Since $\bseta_1$ and $\bseta_2$ are anadromic,
\begin{equation*}
    \bstheta_j = (\bseta_2)_j - (\bseta_1)_j = (\bseta_2)_j - (\bseta_2)_{B-j+1} = -((\bseta_2)_{B-j+1} - (\bseta_1)_j) = -\bstheta_{B-j+1}.
\end{equation*}
In particular, $\argmax_j \bstheta_j = B - (\argmin_j \bstheta_j) + 1$.

(ii) $\bseta(x)_j = (1-x) (\bseta_1)_j + x (\bseta_2)_j = (1-x) (\bseta_1)_{B-j+1} + x (\bseta_2)_{B-j+1} = \bseta(1-x)_{B-j+1}$.

(iii) Let $\bar{\bseta} = (\bseta_1 + \bseta_2)/2$ so that $\bseta(x) = \bar{\bseta} + (x-1/2)\bstheta$ for all $x \in \mathbb{R}$. It is clear that $\sigma(\bseta(x)) \rightarrow \be_{j^+}$ as $x \rightarrow \infty$ since $j^+$ is unique. A similar argument shows that $\sigma(\bseta(x)) \rightarrow \be_{j^-}$ as $x \rightarrow -\infty$.

(iv) Using the expression of $\calI_Z(x)$ in the proof of Lemma \ref{lem:rdp_scalar}, we get
\begin{align*}
    \calI_Z(x) &= \sum_j \bstheta_j^2 \sigma(\bseta(x))_j - \left( \sum_j \bstheta_j \sigma(\bseta(x))_j \right)^2 \\
    &= \sum_j \bstheta_{B-j+1}^2 \sigma(\bseta(1-x))_{B-j+1} - \left( \sum_j \bstheta_{B-j+1} \sigma(\bseta(1-x))_{B-j+1} \right)^2 \quad \text{by (i) and (ii)} \\
    &= \calI_Z(1-x).
\end{align*}

(v) Differentiating $\calI_Z(x)$ and using the above argument gives:
\begin{align*}
    \calI_Z'(x) &= (\bstheta^3)^\top \sigma(\bseta(x)) - 3 \bstheta^\top \sigma(\bseta(x)) (\bstheta^2)^\top \sigma(\bseta(x)) + 2 (\bstheta^\top \sigma(\bseta(x)))^3 \\
    &= -(\bstheta^3)^\top \sigma(\bseta(1-x)) + 3 \bstheta^\top \sigma(\bseta(1-x)) (\bstheta^2)^\top \sigma(\bseta(1-x)) - 2 (\bstheta^\top \sigma(\bseta(1-x)))^3 \\
    &= -\calI_Z'(1-x).
\end{align*}
Then $\calI_Z'(1/2) = -\calI_Z'(1/2)$, so $\calI_Z'(1/2) = 0$ and $x=1/2$ is a stationary point.

(vi) Using the fact that $0 \leq \bstheta_j^2 \leq \bstheta_{j^+}^2$ for all $j$ and
$$\sum_j \bstheta_j \sigma(\bseta(x))_j \geq \bstheta_{j^+} \sigma(\bseta(x))_{j^+} + \bstheta_{j^-} (1 - \sigma(\bseta(x))_{j^+}) = \bstheta_{j^+} \sigma(\bseta(x))_{j^+} - \bstheta_{j^+} (1 - \sigma(\bseta(x))_{j^+}) \geq 0,$$ we have:
\begin{align*}
    \calI_Z(x) &= \sum_j \bstheta_j^2 \sigma(\bseta(x))_j - \left( \sum_j \bstheta_j \sigma(\bseta(x))_j \right)^2 \\
    &\leq \bstheta^2_{j^+} - (\bstheta_{j^+} \sigma(\bseta(x))_{j^+} - \bstheta_{j^+} (1 - \sigma(\bseta(x))_{j^+}))^2 \\
    &= \bstheta_{j^+}^2 (1 - (2 \sigma(\bseta(x))_{j^+} - 1)^2) \\
    &= 4 \bstheta_{j^+}^2 \sigma(\bseta(x))_{j^+} (1 - \sigma(\bseta(x))_{j^+}).
\end{align*}
\end{proof}

\paragraph{Algorithm.} To use Lemma \ref{lem:fisher_anadromic} to compute $M$, we first compute $I^* = \calI_Z(1/2)$ since $x=1/2$ is a stationary point by Lemma \ref{lem:fisher_anadromic}(v). By setting
\begin{equation*}
    4 \bstheta_{j^+}^2 \sigma(\bseta(x))_{j^+} (1 - \sigma(\bseta(x))_{j^+}) \leq I^*
\end{equation*}
and solving this quadratic equation for $\sigma(\bseta(x))_{j^+}$, we can use the bound in Lemma \ref{lem:fisher_anadromic}(vi) to obtain that $\calI_Z(x) \leq I^*$ when $\sigma(\bseta(x))_{j^+} \geq \left(1 + \sqrt{1 - I^* / \bstheta_{j^+}^2} \right) / 2 \geq 1/2$. Since $\sigma(\bseta(x))_{j^+} \rightarrow 1$ as $x \rightarrow \infty$ by (iii), we can determine the value $x_{\max}$ for which $\calI_Z(x) \leq I^*$ when $x \geq x_{\max}$. By (iv), $x_{\min} = 1 - x_{\max}$ satisfies $\calI_Z(x) \leq I^*$ when $x \leq x_{\min}$. We can then do line search in $[x_{\min}, x_{\max}]$ (or equivalently, in $[1/2, x_{\max}]$ by Lemma \ref{lem:fisher_anadromic}(iv)) to obtain $M$.

\section{Choice of Hyperparameters}
\label{sec:hyerparameters}

\begin{table}[h!]
    \begin{subtable}[t]{0.5\textwidth}
    \centering
    \resizebox{\textwidth}{!}{
    \begin{tabular}{ll}
        \toprule
        \textbf{Hyperparameter} & \textbf{Values} \\
        \midrule
        Batch size / cohort size & $600$ \\
        Momentum & $0.5$ \\
        \# Iterations $T$ & $100,200,300,500,1000,2000,3000,5000$ \\
        Step size $\rho$ & $0.001, 0.003, 0.01, 0.03, 0.1, 0.3$ \\
        Gradient norm clip $C$ & $0.25, 0.5, 1, 2, 4, 8$ \\
        Noise multiplier $\sigma$ for Gaussian and SignSGD & $0.5, 1, 2, 3, 5$ \\
        Noise scale $\mu$ for Skellam & $(\sigma C)^2$ \\
        $L_1$-metric DP parameter $\epsilon$ for MVU/I-MVU & $0.25, 0.5, 0.75, 1, 2, 3, 5$ \\
        $\beta$ scaling for I-MVU & 1 \\
        \bottomrule
    \end{tabular}
    }
    \caption{MNIST}
    \end{subtable}%
    \begin{subtable}[t]{0.5\textwidth}
    \centering
    \resizebox{\textwidth}{!}{
    \begin{tabular}{ll}
        \toprule
        \textbf{Hyperparameter} & \textbf{Values} \\
        \midrule
        Batch size / cohort size & $500$ \\
        Momentum & $0.5$ \\
        \# Iterations $T$ & $500,1000,2000,3000,5000,10000,15000$ \\
        Step size $\rho$ & $0.001, 0.003, 0.01, 0.03, 0.1, 0.3$ \\
        Gradient norm clip $C$ & $0.25, 0.5, 1, 2, 4, 8$ \\
        Noise multiplier $\sigma$ for Gaussian and SignSGD & $0.5, 1, 2, 3, 5$ \\
        Noise scale $\mu$ for Skellam & $(\sigma C)^2$ \\
        $L_1$-metric DP parameter $\epsilon$ for MVU/I-MVU & $0.25, 0.5, 0.75, 1, 2, 3, 5$ \\
        $\beta$ scaling for I-MVU & 1 \\
        \bottomrule
    \end{tabular}
    }
    \caption{CIFAR-10}
    \end{subtable}
    \caption{Hyperparameters for $L_2$-bounded client-level DP training
    \label{tab:hyp_l2}}
\end{table}

\begin{table}[h!]
    \centering
    \resizebox{0.5\textwidth}{!}{
    \begin{tabular}{ll}
        \toprule
        \textbf{Hyperparameter} & \textbf{Values} \\
        \midrule
        Batch size / cohort size & $500$ \\
        Momentum & $0.5$ \\
        \# Iterations $T$ & $100,200,300,500,1000,2000,3000,5000$ \\
        Step size $\rho$ & $0.001, 0.003, 0.01, 0.03, 0.1, 0.3$ \\
        Gradient norm clip $C$ & $0.25, 0.5, 1, 2, 4, 8$ \\
        Noise multiplier $\sigma$ for Gaussian and SignSGD & $0.5, 1, 2, 3, 5$ \\
        Noise scale $\mu$ for Skellam & $(\sigma C)^2$ \\
        $L_1$-metric DP parameter $\epsilon$ for MVU/I-MVU & $0.25, 0.5, 0.75, 1, 2, 3, 5$ \\
        $\beta$ scaling for I-MVU & 1 \\
        \bottomrule
    \end{tabular}
    }
    \caption{Hyperparameters for $L_2$-bounded client-level DP CIFAR-10 fine-tuning
    \label{tab:hyp_cifar_finetune}}
\end{table}

\begin{table}[h!]
    \begin{subtable}[t]{0.5\textwidth}
    \centering
    \resizebox{\textwidth}{!}{
    \begin{tabular}{ll}
        \toprule
        \textbf{Hyperparameter} & \textbf{Values} \\
        \midrule
        Batch size / cohort size & $600$ \\
        Momentum & $0.5$ \\
        \# Iterations $T$ & $100,200,300,500,1000,2000,3000,5000$ \\
        Step size $\rho$ & $0.01, 0.03, 0.1, 0.3, 1.0, 3.0$ \\
        Gradient norm clip $C$ & $0.25, 0.5, 1, 2, 4, 8$ \\
        Noise multiplier $\mu$ for Laplace and SignSGD & $0.01, 0.02, 0.03, 0.05, 0.1, 0.2, 0.3, 0.5$ \\
        $L_1$-metric DP parameter $\epsilon$ for I-MVU & $1, 2, 3, 5, 10, 20$ \\
        $\beta$ scaling for I-MVU & 8 \\
        \bottomrule
    \end{tabular}
    }
    \caption{MNIST}
    \end{subtable}%
    \begin{subtable}[t]{0.5\textwidth}
    \centering
    \resizebox{\textwidth}{!}{
    \begin{tabular}{ll}
        \toprule
        \textbf{Hyperparameter} & \textbf{Values} \\
        \midrule
        Batch size / cohort size & $500$ \\
        Momentum & $0.5$ \\
        \# Iterations $T$ & $500,1000,2000,3000,5000,10000,15000$ \\
        Step size $\rho$ & $0.01, 0.03, 0.1, 0.3, 1.0, 3.0$ \\
        Gradient norm clip $C$ & $0.25, 0.5, 1, 2, 4, 8$ \\
        Noise multiplier $\mu$ for Laplace and SignSGD & $0.01, 0.02, 0.03, 0.05, 0.1, 0.2, 0.3, 0.5$ \\
        $L_1$-metric DP parameter $\epsilon$ for I-MVU & $1, 2, 3, 5, 10, 20$ \\
        $\beta$ scaling for I-MVU & 8 \\
        \bottomrule
    \end{tabular}
    }
    \caption{CIFAR-10}
    \end{subtable}
    \caption{Hyperparameters for $L_1$-bounded client-level DP training
    \label{tab:hyp_l1}}
\end{table}

The privacy vs. accuracy curves in the paper plot the Pareto frontier across a grid of hyperparameters.
Table \ref{tab:hyp_l2} gives the range of hyperparameter values used in \autoref{fig:dpsgd_comparison}. Table \ref{tab:hyp_cifar_finetune} gives the range of hyperparameters used in \autoref{fig:dpsgd_cifar_wrn}. Table \ref{tab:hyp_l1} gives the range of hyperparameter values used in \autoref{fig:dpsgd_l1_comparison}.
Tables \ref{tab:femnist_hyperparams_gaussian}, \ref{tab:femnist_hyperparams_imvu} and \ref{tab:femnist_hyperparams_ssgd} give the range of hyperparameter values used in \autoref{fig:dpsgd_femnist}.

\begin{table}[h]
\begin{center}
\begin{tabular}{ll}
\toprule 
Hyperparameter & Values \\ 
\midrule
$\sigma$ for Gaussian & 0.6, 0.8, 1, 2, 4, 6, 8, 10, 16, 20, 24, 32, 40, 50, 60, 64, 70, 80, 90, 100, 110, 120, 128 \\
$\mu$ for Skellam & $(\sigma C)^2$ \\
Server-side learning rate & 0.5, 1, 2 \\
Clipping factor & 0.1, 0.5, 1, 2, 10 \\
\bottomrule
\end{tabular}
\end{center}
\caption{\label{tab:femnist_hyperparams_gaussian}Hyperparameter range for Skellam and Gaussian on FEMNIST.}
\end{table}

\begin{table}[h]
\begin{center}
\begin{tabular}{ll}
\toprule 
Hyperparameter & Values \\ 
\midrule
$\varepsilon$ & 0.25, 0.5, 0.75, 1, 2, 3, 5, 6, 7, 8, 9, 10 \\
Server-side learning rate & 0.5, 1, 2 \\
Scaling factor $\beta$ & 32, 64, 128\\
\bottomrule
\end{tabular}
\end{center}
\caption{\label{tab:femnist_hyperparams_imvu}Hyperparameter range for I-MVU on FEMNIST.}
\end{table}

\begin{table}[h]
\begin{center}
\begin{tabular}{ll}
\toprule 
Hyperparameter & Values \\ 
\midrule
$\sigma$ & 0.6, 0.8, 1, 2, 4, 6, 8, 10, 16, 32, 64, 128 \\
Server-side learning rate & 0.0001, 0.001, 0.01 \\
Clipping factor & 0.5, 1, 2 \\
\bottomrule
\end{tabular}
\end{center}
\caption{\label{tab:femnist_hyperparams_ssgd}Hyperparameter range for SignSGD on FEMNIST.}
\end{table}

\section{Additional Results}

\begin{figure}[t]
    \centering
    \begin{subfigure}{.6\textwidth}
      \centering
      \includegraphics[width=\linewidth]{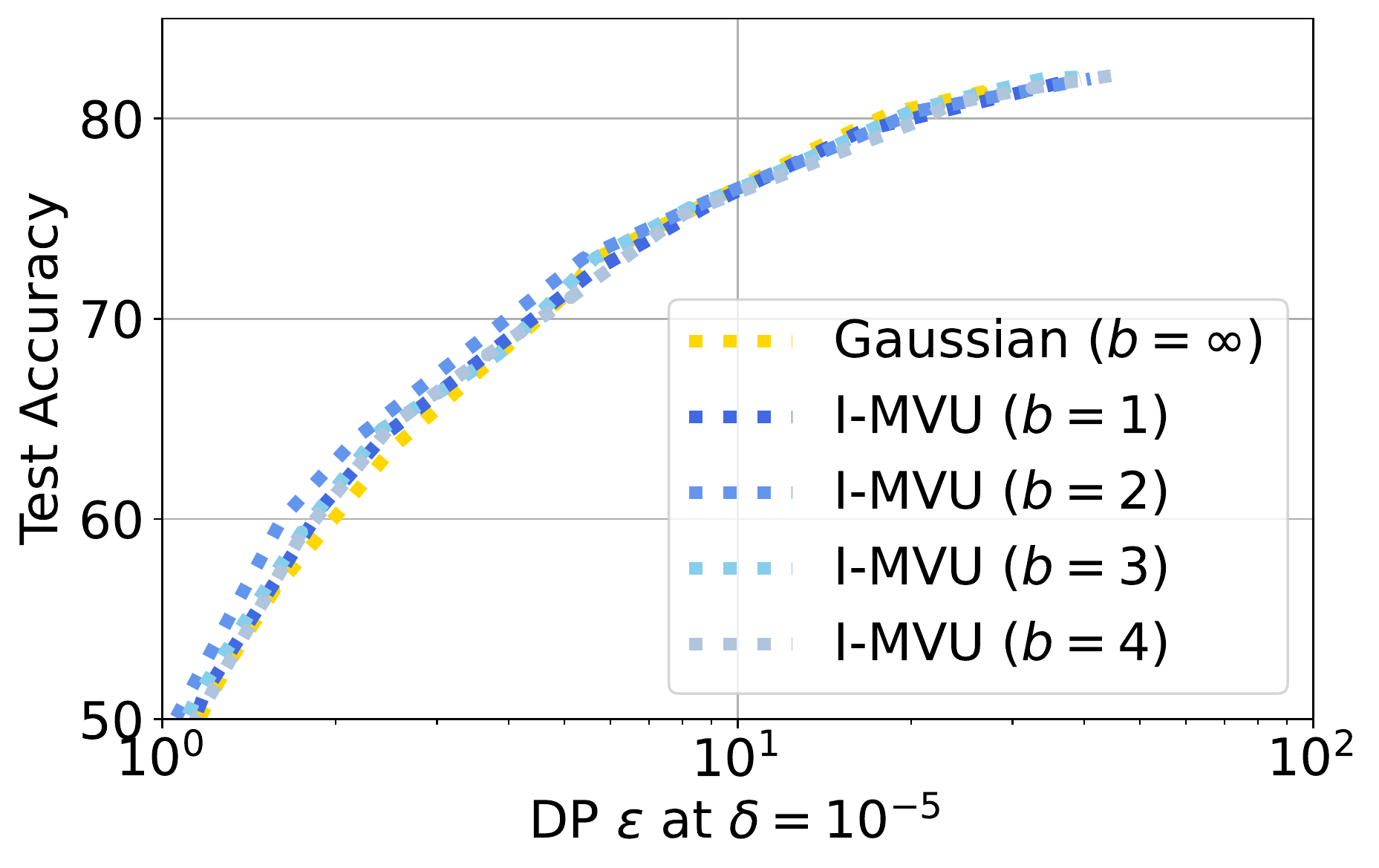}  
    \end{subfigure}
\vspace{-1ex}
\caption{Privacy/accuracy plot for the sample-level DP scenario on FEMNIST. Each point represents a single hyperparameter setting and the Pareto frontier is shown in dashed line. We display the performance of I-MVU with $b$ bits communication budget per coordinate: it is competitive with the non-compressed Gaussian baseline across the entire range of $\epsilon$.}
\label{fig:dpsgd_multiple_bits}
\vspace{-1ex}
\end{figure}

The budget parameter $b$ controls how many bits the I-MVU mechanism outputs. We performed ablation study on $b$ to show that for the sample-level DP experiment in Section \ref{sec:sample_dp}, I-MVU with $b=1$ achieves nearly the same privacy-utility trade-off as with $b=2,3,4$.
Figure \ref{fig:dpsgd_multiple_bits} shows the global model's test accuracy vs. DP $\epsilon$ for the Gaussian mechanism and I-MVU with $b=1,2,3,4$. Evidently, increasing $b$ from 1 to 4 does not result in a significant increase in test accuracy for a fixed value of $\epsilon$. This can be contrasted with the result in \autoref{fig:dpsgd_l1_comparison}, where increasing $b$ from 1 to 4 \emph{does} result in a significant boost in test accuracy.